\documentclass[11pt]{article}
\usepackage{hyperref}
\usepackage{graphicx}
\usepackage{amssymb}
\usepackage[margin=1in]{geometry}
\usepackage{appendix}
\usepackage{setspace}
\spacing{1.5}
\usepackage{titlesec}
\titlespacing*{\section}{0pt}{4pt}{3pt}
\titlespacing*{\subsection}{0pt}{3pt}{2.5pt}

\usepackage{bbm}
\usepackage{algorithm}
\usepackage{algpseudocode}
\usepackage{rotating}
\usepackage{mathtools}
\DeclarePairedDelimiter\ceil{\lceil}{\rceil}

\usepackage{xcolor}
\usepackage{ulem}
\usepackage{colortbl}
\usepackage{tikz}

\usepackage[round]{natbib}
\usepackage{mathtools}

\usepackage{bigstrut}
\usepackage{soul}
\usepackage{booktabs}
\usepackage{multirow}
\usepackage{tabularx}
\usepackage{multicol}
\usepackage{multirow}
\usepackage{amsmath}
\DeclareMathOperator*{\argmax}{arg\,max}
\DeclareMathOperator*{\argmin}{arg\,min}

\DeclareMathOperator{\var}{var}
\DeclareMathOperator{\cov}{cov}

\renewcommand{\emph}[1]{\textit{#1}}
\newcommand{\R}{\mathbb{R}}

\newcommand{\NN}{N}
\newcommand{\x}{\mathbf{x}}

\newcommand{\A}{\mathcal{A}}
\renewcommand{\P}{\mathbf{P}}
\newcommand{\E}{\mathbf{E}}

\newcommand{\K}{\mathcal{K}}
\newcommand{\s}{\mathbf{s}}

\usepackage{accents}

\newtheorem{example}{Example}[section]

\usepackage{natbib}

\bibpunct[, ]{(}{)}{,}{a}{}{,}%

\title{\vspace{-2.5cm}Deep Controlled Learning for Inventory Control}
\date{} 
\author{}

\usepackage{authblk}

\author[1]{Tarkan Temizöz\thanks{Corresponding author: \texttt{t.temizoz@tue.nl}}}
\author[1]{Christina Imdahl}
\author[1]{Remco Dijkman}
\author[1,2]{Douniel Lamghari-Idrissi}
\author[1]{Willem van Jaarsveld}

\affil[1]{Department of Industrial Engineering and Innovation Sciences, Eindhoven University of Technology, PO Box 513, Eindhoven 5600 MB, Netherlands
}
\affil[2]{ASML US LLC, 2625 W Geronimo Pl, Chandler, Arizona 85224, USA}

\makeatletter
\renewenvironment{abstract}{%
  \if@twocolumn
    \section*{\abstractname}%
  \else
    \small
    \begin{center}%
      {\bfseries \abstractname\vspace{\z@}}%
    \end{center}%
    \quotation
  \fi}
  {\if@twocolumn\else\endquotation\fi}
\makeatother
 
\begin{document}
\maketitle

\vspace{-1.25cm}
\begin{abstract}
\noindent The application of Deep Reinforcement Learning (DRL) to inventory management is an emerging field. However, traditional DRL algorithms, originally developed for diverse domains such as game-playing and robotics, may not be well-suited for the specific challenges posed by inventory management. Consequently, these algorithms often fail to outperform established heuristics; for instance, no existing DRL approach consistently surpasses the capped base-stock policy in lost sales inventory control. This highlights a critical gap in the practical application of DRL to inventory management: the highly stochastic nature of inventory problems requires tailored solutions. In response, we propose Deep Controlled Learning (DCL), a new DRL algorithm  designed for highly stochastic problems. DCL is based on approximate policy iteration and incorporates an efficient simulation mechanism, combining Sequential Halving with Common Random Numbers. Our numerical studies demonstrate that DCL consistently outperforms state-of-the-art heuristics and DRL algorithms across various inventory settings, including lost sales, perishable inventory systems, and inventory systems with random lead times. DCL achieves lower average costs in all test cases while maintaining an optimality gap of no more than 0.2\%. Remarkably, this performance is achieved using the same hyperparameter set across all experiments, underscoring the robustness and generalizability of our approach. These findings contribute to the ongoing exploration of tailored DRL algorithms for inventory management, providing a foundation for further research and practical application in this area.
\end{abstract}

\section{Introduction}\label{sec:intro}
Inventory management problems play a critical role in supply chain and logistics as they directly influence the financial performance of a company \citep{Silver2016}. The field encompasses a diverse array of challenges, such as mitigating lost sales due to stockouts \citep[e.g.,][]{Bijvank2023}, managing products with limited shelf life \citep[e.g.,][]{karaesmen2011managing}, and inventory control with random replenishment lead times \citep[e.g.][Chapter 7]{zipkin2000foundations}. The decision-making process in these systems is complicated due to the high degree of stochasticity resulting from exogenous factors such as demand uncertainty and stochastic lead times. In addition, each of these problems comes with unique operational dynamics, making their solutions substantially distinct from one another. For instance, managing perishable products may necessitate inventory reduction strategies to minimize waste \citep{nahmias2011perishable}, while navigating uncertain lead times may demand maintaining higher stock levels and robust contingency plans \citep[][Chapter 7]{zipkin2000foundations}. As a result, addressing these challenges requires tailored approaches.

Optimal solutions for these problems are often computationally intractable \citep{boute2021deep} due to the exponential growth in possible combinations of states, actions, and uncertain scenarios, resulting from their high dimensionality and stochastic nature. Consequently, researchers resort to approximate methods to model and solve these problems \citep[][Chapter 6]{zipkin2000foundations}. In this context, Markov Decision Processes (MDPs) are frequently employed to model inventory management problems, as they capture underlying uncertainties and accommodate sequential decision-making processes. To address the intractability of solving MDPs, various approximate dynamic programming (ADP) methods have been explored \citep{powell2011approximate}. However, these methods often have problem-specific limitations, such as requiring a maximum lead time of one \citep[e.g.][]{Chen2014adp}, or depend on exploiting cost-structure information, such as the cost function exhibiting $L^\natural$ convexity \citep[e.g.][]{sun2016quadratic}. However, as \citet{Cachon2020} notes, achieving more impactful inventory management requires more broadly applicable approaches.

Deep reinforcement learning (DRL) offers a promising alternative by integrating neural networks with reinforcement learning, enabling the learning of complex decision-making processes in high-dimensional state spaces \citep{mnih2015human}. DRL algorithms utilize neural networks to approximate the relationship between system states and decisions, iteratively refining the decision-making process. DRL has the potential to overcome some limitations of traditional methods, including problem-specific constraints, as shown by its empirical success as a general-purpose framework across a wide range of applications, including Atari games \citep{mnih2015human} and board games \citep{silver2018general}. Building on these successes, DRL has been applied to inventory management problems, achieving results comparable to heuristic policies \citep{gijsbrechts2019can, oroojlooyjadid2021deep, vanvuchelenPPO}.

Despite their successes, DRL algorithms employed in prior studies, such as Asynchronous Advantage Actor-Critic (A3C) \citep{mnih2016asynchronous}, Deep Q-Learning (DQN) \citep{mnih2015human}, and Proximal Policy Optimization (PPO) \citep{schulman2017proximal}, are not explicitly designed to handle the high level of stochasticity of the MDPs commonly encountered in inventory management problems \citep{trimponias2023reinforcement}. As highlighted by \citet{powell2020reinforcement}, such MDPs, also known as Input-Driven MDPs \citep{mao2018variance} or MDPs with exogenous variables \citep{dietterich2018discovering}, are characterized by an independent stochasticity source exogenous to the MDP that is uncontrollable by decisions. A typical example is \emph{customer demand}, which is independent of replenishment orders in many inventory management applications.

Consequently, these algorithms may not consistently outperform existing state-of-the-art heuristics in inventory management, as demonstrated by their performance in the lost sales problem \citep{gijsbrechts2019can} and perishable inventory systems \citep{bram2022}. This limitation can be attributed to several factors that render these algorithms less suitable for addressing such problems, preventing them from being used in practice. First, algorithms such as A3C and PPO update the policy using trajectories generated by the current policy, which often results in each state being visited a limited number of times. Limited state revisitation and significant variations in trajectory costs, resulting from the high stochasticity of these problems, make it difficult to accurately assess the relative values of actions in a state. Evaluating states under multiple exogenous scenarios is essential to address this issue \citep{dietterich2018discovering}. However, generating multiple scenarios for state evaluation can be computationally expensive and inefficient, necessitating approaches to address this limitation. Moreover, all three algorithms utilize neural networks to approximate costs tied to policies. In inventory management, stochastic exogenous inputs, such as uncertain demand or lead times, can introduce significant noise into these approximations. This noise can then degrade the accuracy of the estimate of expected trajectory costs, thereby affecting the efficacy of derived policies \citep[see][for a similar discussion]{Lazaric}. In summary, there is a need for DRL algorithms designed to handle these challenges and cater to the requirements of inventory management problems.

This paper proposes Deep Controlled Learning (DCL), a new end-to-end DRL algorithm tailored for inventory management applications. DCL is an approximate policy iteration algorithm that iteratively improves policies by casting reinforcement learning as a classification problem. It employs simulations to collect state-action pairs to form a dataset, which is then used to train neural networks for policy representation. For each state in the dataset, an estimated optimal action is determined by revisiting and evaluating that state under multiple exogenous scenarios and selecting the action with the least estimated expected costs over a trajectory. This estimated action serves as the label for the corresponding state in the classification task, guiding the neural network to map that state to this action. In doing so, DCL iteratively forms datasets and refines policies. 

From a methodological perspective, DCL aligns with a category of algorithms known as Classification Based Policy Iteration (CBPI) \citep[e.g.,][]{lagoudakis2003reinforcement, Lazaric}. Unlike other DRL algorithms such as PPO and A3C, which utilize the 'policy gradient' formula to adjust the probabilities of selecting each action based on their estimated advantages, CBPI treats actions explicitly as class labels. This allows for direct classification of actions, potentially leading to more stable learning progress. Furthermore, the CBPI framework avoids the need for approximating costs via any function approximator. Instead, it directly estimates optimal actions through simulations. Consequently, DCL relies on neural networks solely for policy representation, circumventing their use in cost approximation, which can introduce high levels of noise into expected cost estimates. This design choice helps DCL to sidestep issues associated with inaccurate cost approximations, focusing instead on robust policy representation.

Moreover, DCL implements several strategies to reduce the computational burden of simulations and enhance performance, hence extending the CBPI framework. DCL uses a variance control mechanism called Common Random Numbers (CRN) \citep[see][]{law2000simulation}, which helps decrease variance and improve the accuracy of determining the labeled action for a state. This is achieved utilizing the same exogenous scenarios for evaluating each action. This approach ensures that the different actions are assessed under the same conditions, thus reducing the variance in the comparison. In addition, DCL integrates a state-of-the-art bandit algorithm, Sequential Halving (SH) \citep{seqhalving}, to efficiently allocate exogenous scenarios to actions that are more promising to be optimal. This is in contrast to the conventional approach that uniformly distributes simulation resources to estimate the optimal action for a given state, used by previous CBPI applications. To our knowledge, DCL is the first work that combines and integrates SH with CRN for optimal action estimation within the DRL framework. This combination of techniques enables DCL to address high stochasticity more effectively. 

We evaluate DCL on three well-known inventory problems: lost sales inventory control, perishable inventory systems, and inventory systems with random lead times. Our results demonstrate that DCL consistently outperforms existing state-of-the-art heuristics and achieves an optimality gap of at most 0.2\% for each of these problems, a significant improvement over the 3-6\% gap observed with A3C in the lost sales problem \citep{gijsbrechts2019can}. Remarkably, the same set of hyperparameters is used across all experiments, demonstrating the robustness and generalizability of our proposed approach. Furthermore, we find that the combination of SH and CRN together reduces simulation effort by about two orders of magnitude. This work, therefore, represents a significant advancement in developing a DRL algorithm for inventory management problems, providing efficient solutions for a wide range of challenging tasks. A key insight arising from this paper is that a single general-purpose algorithm may outperform the best heuristic policies for three difficult inventory problems, which is of interest to practitioners as well as researchers.

This paper is organized as follows: \S\ref{sec:literature} reviews the literature, \S\ref{sec:mdp} provides the details of Markov Decision Processes with Exogenous Inputs, and \S\ref{sec:algorithm} introduces the DCL algorithm. Experimental setup and numerical results are presented in \S\ref{sec:experimentsetup} and \S\ref{sec:results}, while \S\ref{sec:discussion} concludes the paper.

\section{Literature Review}\label{sec:literature}

We first discuss well-known heuristic inventory policies for three difficult inventory problems considered in this paper. We then delve into the emerging use of DRL in inventory management. 

\subsection{Heuristic Policies for three Difficult Inventory Problems}\label{sec:inventoryliterature}

Many leading heuristics for the inventory problems considered in this paper are modifications of the \emph{base-stock policy} (BSP), which is prevalent in a wide range of inventory systems due to its simplicity \citep{Arrow1958}. It places an order each period to increase the \emph{inventory position}, which equals inventory on-hand plus inventory in pipeline minus back-orders, to a fixed number known as the \emph{base-stock level}. Its simplicity stems from having only this single parameter, which can either be efficiently optimized or approximated, depending on the model. 

\textbf{Lost sales inventory control} is a canonical inventory problem, and its complex stochastic nature has prompted extensive research \citep[see][]{Bijvank2023}. Finding the optimal policy for lost sales systems with lead times is impractical because it depends on each outstanding order \citep{morton1969bounds,zipkin2008old}. Prior studies have established the asymptotic optimality of simple policies, such as base-stock and constant order policies, as either the penalty cost or the lead-time grows large \citep{huh2009asymptotic, goldberg2016asymptotic, xin2016optimality}, but finding optimal policies beyond these regimes remains intractable even for moderately long lead times. Therefore, researchers and practitioners often resort to heuristic approaches. \citet{zipkin2008old} offers a collection of benchmark problems for heuristic evaluation, highlighting the well-performing \emph{myopic} policies \citep{morton1971near} in these benchmarks. \citet{Xin2021} studies the \emph{capped base-stock} policy, first introduced by \citet{Johansen2008}, building upon the asymptotic properties of prior approaches. 
It works akin to BSP by ordering up to the base-stock level, but it also introduces a cap on the maximum number of orders.


\textbf{Perishable inventory systems} with a fixed product lifetime have triggered much inventory research throughout the years, as shown in comprehensive reviews \citep{nahmias1975comparison, karaesmen2011managing, nahmias2011perishable, Chao2018}. In these systems, any remaining inventory that reaches the end of its lifetime perishes and is removed with accompanying waste costs. Finding the optimal policy for these systems is intractable as it necessitates tracking the inventory of each age group due to the fixed product lifetime, resulting in a multidimensional state space \citep{nahmias1975optimal}. Additionally, one needs to consider the \emph{issuance policy}—the order in which items are sold. While the first-in-first-out (FIFO) policy prioritizes selling the oldest items first, the last-in-first-out (LIFO) policy does the opposite by selling the youngest items first. Interestingly, studies focusing on systems under the LIFO policy are sparse \citep[e.g.,][]{cohen1978lifo, bu2023} despite their common occurrence in retail operations \citep{minner2010}. Also, BSP falls short by considering only the inventory position, failing to account for the age of units in stock and transit.

To fill this gap, \citet{Haijema2019} introduce \emph{BSP-low-EW}—a modified BSP that leverages the concept of estimated waste, or the expected number of units in inventory that will perish during lead time \citep{broekmeulen2009heuristic}. The policy utilizes two order-up-to levels based on the inventory position and factors in the estimated waste during lead time, under the assumption that demand per period is deterministic and equals the mean. As shown by a large-scale simulation study across various settings, including different issuing policies, BSP-low-EW emerges as the top-performing heuristic.

\textbf{Inventory systems with random lead times} pose significant challenges, particularly when orders may \emph{cross}, which happens when an order that is placed after another order is received before that order. In the presence of order-crossing, states must include the entire order history \citep{Muthuraman2015}, and consequently, the traditional BSP is no longer optimal \citep{Disney2016}. Yet, heuristic solutions tailored to address these conditions remain scarce \citep{Stolyar2022}, and typically the system is optimized within the framework of base-stock or $(r, Q)$ policies \citep{Bradley2005,Ang2017}. \citet{Stolyar2022} propose the \emph{generalized base-stock} policy, which sets a dynamic target for the inventory in the pipeline that is adjusted based on the inventory level. This enables the policy to differentiate between the inventory level and the pipeline inventory - a crucial distinction when lead times are stochastic. The generalized base-stock policy demonstrates significant cost savings compared to the BSP.

\subsection{Deep Reinforcement Learning for Inventory Management}

The DRL literature has seen several groundbreaking algorithms, such as DQN \citep{mnih2015human}, A3C \citep{mnih2016asynchronous}, PPO \citep{schulman2017proximal} and AlphaZero \citep{silver2018general}, demonstrating remarkable performance in diverse domains, including early applications to inventory problems \citep[see][]{boute2021deep}.

The potential of DRL for inventory problems is first identified by early work including \cite{oroojlooyjadid2021deep}, who apply DQN to the beer game, and \cite{gijsbrechts2019can}, who apply A3C to three classic inventory problems: lost sales, dual sourcing, and multi-echelon inventory control. With extensive hyperparameter tuning, A3C demonstrates competitive performance against state-of-the-art heuristics, indicating the promise of DRL as a general-purpose technology in this domain. While being outperformed by the capped base-stock and myopic-2 (an extension of the myopic policy) policies for the lost sales problem, the authors find that A3C performs better than BSP and obtains a 3-6\% optimality gap for instances with a computationally feasible optimal policy. Moreover, \citet{bram2022} incorporate potential-based reward shaping in the DQN algorithm for perishable inventory systems, showing that shaping from a teacher policy can improve the learning process of a DRL algorithm when a good heuristic like BSP-low-EW is present. However, their shaped-DQN algorithm cannot match the performance of BSP-low-EW in some settings, particularly when the lifetime of a product is large. Other studies specifically deploy PPO to address inventory challenges, exhibiting varying degrees of performance efficacy \citep[e.g.,][]{vanvuchelenPPO,vanHezewijk2023, geevers2023multiechelon, KAYNOV2024109088, DEHAYBE2024433, WANG2024109133} while \citet{Stranieri2024} compares the performances of various DRL algorithms in two-echelon inventory control systems.

Our research distinguishes itself from these listed applications of DRL to inventory management, not merely by applying pre-existing DRL algorithms but by architecting a new algorithm tailored for such problems. In this vein, AlphaZero provides a relevant comparative context. \citet{silver2018general} introduced AlphaZero as a general solution for deterministic two-player board games, capitalizing on the known dynamics and rules governing these games. Analogously, our approach, DCL, proposes a general algorithm for environments impacted by stochastic exogenous inputs featuring a single decision-maker. This method reflects a collective effort to enhance the fit between the algorithm design and the characteristics of inventory management.

\section{Markov Decision Processes with Exogeneous Inputs}\label{sec:mdp}
We explore a specialized class of Markov Decision Processes that will be referred to as \emph{MDPs with Exogenous Inputs} (MDP-EI). This framework can be particularly useful in various operational settings, including inventory management problems, where independent stochastic sources such as demand and lead time significantly influence the decision-making process. We can model these stochastic sources as exogenous inputs to the MDP \citep[cf. the unified framework in][]{powell2019unified}. Consequently, our proposed DCL algorithm adopts MDP-EI to model inventory management problems.

\begin{figure}[ht]
\centering
{\includegraphics{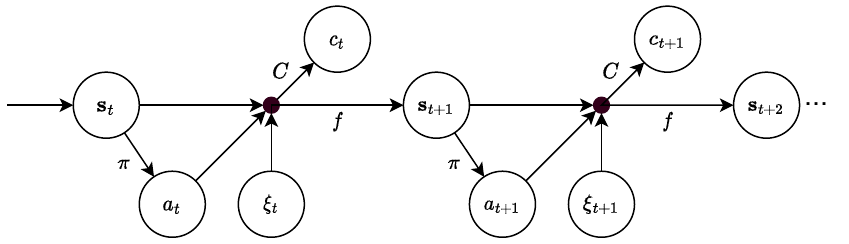}}
\caption{Graphical representation of the MDP-EI, with arrows indicating direct dependence.}\label{fig:mdp}
\end{figure}

In the MDP-EI model, we consider a finite discrete state space $\mathcal{S}$, a finite action space $\A = \{0, 1, \ldots, m\}$, and a set of possible exogenous inputs $\mathcal{W}$. For any given state $\s \in \mathcal{S}$, we represent the applicable actions as $A_\s \subseteq \A$. A deterministic policy, denoted by $\pi$, is a function $\pi: \mathcal{S} \to \A$ that assigns to each state $\s$ a valid action $\pi(\s) \in \A_\s$. In each period $t$, the next state $\s_{t+1}$ and the incurred cost $c_t$ are determined by a combination of the current state $\s_t$, the action $a_t$ taken, and the exogenous input $\xi_t$. Here, $\xi_t \in \mathcal{W}$ represents an exogenous stochastic input of any dimension observed after the execution of action $a_t$ at period $t$. These inputs are assumed to be independent of the states and actions. For ease of exposition, we shall also assume that the inputs are independently and identically distributed, which is a common assumption in inventory management literature \citep[][Chapter 6]{zipkin2000foundations} and in practical settings \citep[e.g.,][]{Houtum2015spare}. Consequently, following the policy $\pi$ from state $\s$ results in a random trajectory composed of states, actions, exogenous inputs, and costs. This trajectory, representing the evolution of the system, can be described by the sequence $(\s_0, a_0, \xi_0, c_0, \s_1, a_1, \xi_1, c_1, \s_{2}, a_{2}, \xi_2, c_2, \s_3, \dots)$, where $\s_0 = \s$ and $a_t = \pi(\s_t)$. See Figure \ref{fig:mdp} for the visual depiction of this evolution.

\newcommand{\Input}{\mathbf{\Xi}}

Accordingly, we define the MDP-EI as a tuple $\mathcal{M} = \langle \mathcal{S}, \A, \mathcal{W}, \Input, f, C, \alpha, \s_{0} \rangle$. Here, $\mathcal{S}$ and $\A$ denote the finite sets of states and actions, respectively, while $\mathcal{W}$ is the set of exogenous inputs. The variable $\Input$ represents a generic random variable such that $\xi_t\sim \Input$. We shall assume that $\Input$ is distributed such that realizations $\xi_t\sim\Input$ can be generated efficiently. State transitions are determined by the function $f: \mathcal{S} \times \A \times \mathcal{W} \to \mathcal{S}$. This function maps a state-action-input tuple $(\s, a, \xi)$ to a new state $\s'=f(\s,a,\xi)$. The cost function $C: \mathcal{S} \times \A \times \mathcal{W} \to \mathbb{R}$ assigns a cost $c = C(\s,a,\xi)$ to each state-action-input combination. Based on this formalization, we can derive the state transition probability and the expected cost when executing action $a$ in state $\s$ as follows:
\begin{align}
&\P(\s'|\s,a) = \E_{\xi \sim \Input}[\mathbbm{1}_{[\s'=f(\s,a,\xi)]}], &\s,\s'\in\mathcal{S}, a\in \A_{\s}\label{eq:exogenous1}, \\
&\bar{C}(\s,a) = \E_{\xi \sim \Input}[C(\s,a,\xi)], & \s\in \mathcal{S}, a\in \A_{\s}. \label{eq:exogenous2}
\end{align}
In the MDP-EI model, costs accumulate over an infinite horizon. Let $\alpha\in (0,1]$ denote the discount factor. For $\alpha<1$, costs are discounted, while $\alpha=1$ corresponds to the average cost criterion. We assume that the system starts from an initial state $\s_{0}$, which could be deterministic or random with an arbitrary distribution, depending on the specific problem context. Appendix~\ref{app:invmodelformulation} illustrates these concepts for three inventory problems considered in this paper.

\section{Algorithm}\label{sec:algorithm}
In \S\ref{sec:API}, we discuss policy iteration and approximate policy iteration. In \S\ref{sec:dcl}, we motivate and discuss DCL and explain the key ideas underlying its development.
\subsection{Approximate Policy Iteration}\label{sec:API}
\textbf{Policy iteration}, a fundamental concept in MDPs, is an iterative method that continually evaluates and improves a given policy to find the optimal solution. While we discuss it next within the context of the MDP-EI modeling where $\alpha<1$, it is important to note that the method is not limited to $\alpha<1$ and the approximations are also valid for average cost problems. In MDP-EI, following a policy $\pi$ from a state $\s$ results in a random trajectory. The discounted costs accumulated over such a trajectory, with an initial state $\s_0=\s$, can be expressed as follows:
\begin{align}
V_\pi(\s) := \sum_{t=0}^{\infty}\alpha^{t}c_t, \text{ with } a_{t} = \pi(\s_{t}),\text{ } \xi_t \sim \Input,\text{ } c_t = C(\s_t,a_t,\xi_t), \text{ and } \s_{t+1} = f(\s_t,a_t,\xi_t). \label{eq:valuefnc}
\end{align}

The value function $v_{\pi}(\s)$ for policy $\pi$ represents the expected value of $V_\pi(\cdot)$ over the distribution of the exogenous inputs $\xi$, that is, $v_\pi(\s)=\E_\xi[V_\pi(\s)]$. This function expresses the expected discounted costs over an infinite horizon, starting from state $\s$ and following policy $\pi$, which satisfies \citep[see][for details]{puterman2014markov}:
\begin{align}
v_\pi(\s)= \bar{C}(\s,\pi(\s)) +\alpha \sum_{\s'\in\mathcal{S}} \P(\s'|\s,\pi(\s)) v_\pi(\s'),\text{ }\forall\s\in \mathcal{S}. \label{eq:expvaluefnc}
\end{align}

While the value function provides an overall measure of policy performance, assessing the value of individual actions can be useful. The action-value function $q_\pi(\s,a)$ represents the expected value of $Q_\pi(\s,a)$, where $Q_\pi(\s,a)$ denotes the costs incurred over a trajectory with initial state $\s_0=\s$ and initial action $a_0=a$:
\begin{align}
Q_\pi(\s,a) := \sum_{t=0}^{\infty}\alpha^{t}c_t, \text{ with } a_{t, t>0} = \pi(\s_{t}),\text{ } \xi_t \sim \Input,\text{ } c_t = C(\s_t,a_t,\xi_t), \text{ and } \s_{t+1} = f(\s_t,a_t,\xi_t). \label{eq:actvaluefnc}
\end{align}
Hence, the action-value function $q_\pi(\s,a)$ can be expressed as \citep[see][for details]{puterman2014markov}:
\begin{align}
q_\pi(\s,a) = \bar{C}(\s,a) +\alpha \sum_{\s'\in\mathcal{S}}
\P(\s'|\s,a) v_\pi(\s'), \text{ }\forall \s \in\mathcal{S}, \text{ }  \forall a \in \A_\s. \label{eq:expactvaluefnc}
\end{align}

An improved policy over  $\pi$, denoted as $\pi^+(\cdot)$, is one that minimizes $q_\pi(\s,a)$ for all possible states $\s$:
\begin{align}
    \pi^+(\s):=\argmin_{a\in \A_\s} q_\pi(\s,a),\text{ }\text{ } \forall \s \in\mathcal{S}\label{eq:exactpolicyiteration}
\end{align} 

Note that $\pi^+(\s)$ indicates the optimal action to take in state $\s$ before adopting policy $\pi$ for subsequent states. In cases of ties in the argmin operation, we can break them arbitrarily. Repeated improvement of a policy in this manner is known as \emph{policy iteration} and leads to the optimal policy under a broad range of assumptions \citep{puterman2014markov}. Nonetheless, this approach requires the calculation of the action-value function $q_\pi(\s,a)$ for each state-action pair $(\s,a), \text{ }\s \in \mathcal{S}, \text{ }a \in \mathcal{A}$, which involves resolving the system of equations \eqref{eq:expvaluefnc}. The latter task can become computationally challenging when $\mathcal{S}$ is large, creating a demand for more efficient strategies.

\textbf{Approximate policy iteration} strategies address the computational burden of exact policy iteration by estimating the action-value function and approximating the improved policy. The goal is to avoid solving the system of equations \eqref{eq:expvaluefnc} and to refrain from estimating the action-value function for every possible state-action pair, respectively. These adjustments make approximate policy iteration tractable for large state spaces. 

A prominent approach for estimating the action-value function for a given state-action pair $(\s,a)$ is the application of \emph{rollout simulations}, a method first proposed by \citet{Tesauro}. Within the context of MDP-EI, rollout simulations facilitate the generation of numerous independent replications of the trajectory cost \eqref{eq:actvaluefnc}, each under a unique exogenous scenario. The empirical mean of these replications offers an estimate of the action-value function, a process we referred to as \emph{evaluating states under multiple exogenous scenarios} in \S\ref{sec:intro}. 

In order to facilitate these simulations and obtain realizations of the trajectory costs, we require an approximation approach, given that MDP-EI are modeled with an infinite horizon (see \S\ref{sec:mdp}). To do so, we truncate the trajectory length after $H$ steps, thereby setting a finite horizon. Given the distribution of exogenous inputs $\Input$, we can generate an exogenous scenario $\boldsymbol\xi$ of length $H$ as follows:
\begin{align}
\boldsymbol\xi=\{\xi_0, \xi_1, \dots, \xi_{H-1} \}, \text{ with } \forall t\in \{0,\ldots,H-1\}: \text{ } \xi_t \sim \Input. \label{eq:exoscenario}
\end{align} 
We can then approximate $Q_\pi(\s,a)$ by initiating the trajectory in state $\s_0=\s$, taking action $a_0=a$, and summing the discounted costs over the finite horizon and under the exogenous scenario $\boldsymbol\xi$:
\begin{align}
\hat{Q}_\pi(\s,a|\boldsymbol\xi):= \sum_{t=0}^{H-1} \alpha^{t} c_t, \text{ with } a_{t, t>0}=\pi(\s_{t}), \text{ } c_t=C(\s_t,a_t,\xi_t),\text{ and }\s_{t+1}=f(\s_t,a_t,\xi_t). \label{eq:qvalestimate} 
\end{align}

The approximated trajectory costs for a state-action pair often exhibit considerable variance due to the diverse exogenous scenarios. Mitigating this variability and improving the precision of action-value function estimation require generating numerous exogenous scenarios, each yielding a distinct approximate trajectory cost. Inadequate estimation of the action-value functions can hinder policy improvement. The potential pitfalls of this practice and the benefits of averaging multiple scenarios are evident in Example \ref{ex1:api} in Appendix \ref{app:examples}.

Suppose we generate $M$ independent rollout simulations to estimate the action-value function. The resulting estimate of the action-value function for state $\s$ satisfies:
\begin{align}
\hat{q}_{\pi}(\s,a)=\sum_{r=1}^{M} \frac{1}{M}\hat{Q}_{\pi}(\s,a|\boldsymbol\xi_{r}), \text{ }  \forall a \in \A_\s.\label{eq:expqvalestimate}
\end{align}

Subsequently, the improved action for state $\s$ before following policy $\pi$ can then be estimated as $\hat{\pi}^{+}(\s) := \argmin_{a \in \A_s}\hat{q}_{\pi}(\s,a)$, where $\hat{\pi}^{+}(\s)$ is referred to as the \emph{simulation-based action} for state $\s$ (cf. \eqref{eq:exactpolicyiteration}).  However, obtaining the simulation-based action for each state in $\mathcal{S}$ is computationally expensive for large state spaces. To circumvent this, we sample $N$ states from $\mathcal{S}$ and compute the corresponding simulation-based action for each sampled state.

Still, the simulation-based policy, denoted as $\hat{\pi}^+(\cdot)$, covers only the sampled states. To extend the search to the entire state-space $\mathcal{S}$, a policy approximation is required that can offer a policy representation applicable to all states \citep{boute2021deep}. 
The policy approximation problem is formulated as a classification task, where a classification model parameterized by $\theta$ is trained to map states to actions, effectively approximating the policy $\hat{\pi}^+(\cdot)$. The resulting policy, $\pi_\theta$, serves as a viable approximation of the policy $\hat{\pi}^+(\cdot)$ for all states in $\mathcal{S}$; however, it is not necessarily an improvement over policy $\pi$ due to potential approximation errors. Such errors are particularly likely when the parameters $H$, $M$ or $N$ are set too low, resulting in significant approximation errors.

The practical application of the policy improvement process and the subsequent policy approximation requires careful consideration of computational complexity and approximation accuracy, which hinge on parameters $H$, $M$, and $N$. We next delve deeper into DCL and explain its strategies to overcome practical challenges.  

\subsection{Deep Controlled Learning}\label{sec:dcl}
While \citet{Lazaric} demonstrate that increasing any of $H$ (the horizon length), $M$ (the number of exogenous scenarios per state-action pair), and $N$ (the number of states to be sampled) theoretically enhances the policy improvement process, DCL algorithm focuses on optimizing its practical performance with fixed $H, M$, and $N$ for solving MDP-EI. In this way, DCL seeks to deliver higher performance levels for given computational resources, thereby increasing efficiency.

Algorithm \ref{alg:DCL} provides a pseudo-code of the DCL algorithm. Given an MDP-EI model $\mathcal{M}$, an initial policy $\pi_0$, and a selection of hyperparameters, including $H$, $M$, and $N$, DCL executes $n$ steps of approximate policy improvement, indexed by $i=0,1,\ldots, n-1$. This procedure iteratively refines the initial policy. During each iteration $i$, DCL begins by gathering $N$ data samples $\K_i=\{(\s_k,\hat{\pi}_i^+(\s_k))|k\in \{1,\ldots,N\}\}$. The collection of samples constitutes states and their corresponding simulation-based actions. To perform this process efficiently, DCL leverages the available computing power by parallelizing the simulation process, which involves selecting states for the dataset and determining their simulation-based actions. This is achieved by uniformly distributing the task of sampling state-action pairs among the available threads. Each thread is responsible for sampling $\lceil N / w \rceil$ state-action pairs, where $N$ is the total number of samples to be collected and $w$ denotes the number of threads (as shown on Line \ref{alg1:paralel} in Algorithm \ref{alg:DCL}). Subsequently, DCL generates the policy $\pi_{i+1}$ by training a neural network \emph{classifier}, i.e., a neural network that can prescribe an action for any state. This classifier works to approximate the simulation-based policy reflected by the dataset $\K_i$.

\begin{algorithm}
\caption{Deep Controlled Learning}\label{alg:DCL}
\begin{algorithmic}[1]
\State \textbf{Input}: MDP model: $\mathcal{M} = \langle \mathcal{S}, \A, \mathcal{W}, \Input, f, C, \alpha, \s_{0} \rangle$, initial policy:  $\pi_{0}$, neural network structure: $\NN_{\theta}$, number of approximate policy iterations: $n$, number of states to be collected: $N$, number of exogenous scenarios per state-action pairs: $M$, depth of the rollouts (horizon length): $H$, length of the warmup period: $L$, number of threads: $w$
\For{$i=0, 1, \dots, n-1 $}
\State $\K_{i} = \{\} $, the dataset
\For{\textbf{each} $thread=1.\dots,w$}  \textbf{in parallel}  \label{alg1:paralel}
\algorithmiccomment{parallelization}
\State $\s_1 = SampleStartState(\mathcal{M}, \pi_i, L)$ \label{alg1:s1} \algorithmiccomment{sample starting state}
\State Generate exogenous scenario $\boldsymbol\xi$ by \eqref{eq:exoscenario}, $|\boldsymbol\xi| = \lceil{N / w}\rceil$  \label{alg1:exothread}
\For{$k=1, \dots, \lceil{N / w}\rceil$} \label{alg1:samp1}
\State Find $\hat{\pi}_i^+(\s_k) = Simulator(\mathcal{M}, \s_{k},\pi_{i}, M, H)$  \algorithmiccomment{SH with CRN} \label{alg1:seqhal}
\State Add ($\s_{k}$, $\hat{\pi}_i^+(\s_k)$) to the data set $\K_{i}$ \label{alg1:samp2}
\State $\s_{k+1} = f(\s_{k}, \hat{\pi}_i^+(\s_k), \xi_{k-1})$ \label{alg1:sampnew}
\EndFor
\EndFor
\State $\pi_{i+1} = Classifier(N_{\theta}, \K_i)$
\algorithmiccomment{training neural networks}  \label{alg1:classifier}
\EndFor 
\State \textbf{Output}: $\pi_{1}, \dots, \pi_{n}$ 
\end{algorithmic}
\end{algorithm}

In the remainder of this section, we first demonstrate how DCL samples states for inclusion in the data set $\K$. Next, we explain the use of the SH algorithm with CRN to determine the simulation-based action for each sampled state. Finally, we describe the application of neural networks as a classification algorithm for policy representation.

\noindent \textbf{State sampling}

To ensure comprehensive coverage of the state space and develop a robust policy relevant to various scenarios and states \citep{sutton2018reinforcement}, we first explain the rationale behind how each thread samples its first state, $s_1$, for inclusion in the dataset. Starting from an initial state $s_{0}$ specified by the MDP-EI model $\mathcal{M}$, each thread generates an exogenous scenario $\boldsymbol{\xi}$ of length $L$ (the length of the warm-up period, a hyperparameter of the algorithm). Following the current policy $\pi_i$ through this trajectory allows the exploration of diverse state transitions. The state observed at the end of the trajectory becomes the starting state (as depicted in Algorithm \ref{alg:samplestate}, invoked on Line \ref{alg1:s1} in Algorithm \ref{alg:DCL}). This state is then incorporated into the dataset after determining its simulation-based action, ensuring that each thread starts from a distinct state.

\begin{algorithm}
\caption{SampleStartState}\label{alg:samplestate}
\begin{algorithmic}[1]
\State \textbf{Input}: $\mathcal{M}, \pi, L$
\State  Generate an exogenous scenario $\boldsymbol\xi$ by \eqref{eq:exoscenario}, $|\boldsymbol\xi| = L$ 
\For{$j=0, \dots, L-1$}
\State $\s_{j+1} = f(\s_{j}, \pi(\s_j), \xi_{j})$
\EndFor
\State \textbf{Output}: $\s_{L}$ 
\end{algorithmic}
\end{algorithm}

The creation of the remaining states for $\K_i$ is influenced by the simulation-based policy $\hat{\pi}_i^+(\cdot)$, rather than the current policy $\pi_i$. Each thread generates and retains an exogenous scenario $\boldsymbol{\xi}$, transitioning to a new state once the simulation-based action for the sampled state has been identified (as shown on Lines \ref{alg1:exothread} and \ref{alg1:sampnew} in Algorithm \ref{alg:DCL}). This strategy ensures that the policy improvement is meaningful and applies to states that the system is likely to visit in practice \citep{Ernst2005}. In other words, it helps ensure that the improved policy is not merely theoretically effective but also practically relevant to the actual operation of the system.

\noindent \textbf{Identifying simulation-based actions}

Once a starting state $\s$ for the data set is selected, DCL identifies the corresponding simulation-based action $\hat{\pi}^+(\s)$ by finding the action associated with the lowest estimate of the action-value function \eqref{eq:expqvalestimate}. In performing this task, DCL maintains a fixed budget of exogenous scenarios, denoted as $B_\s = M |\A_{\s}|$, where $B_\s$ represents the total rollout budget for state $\s$. Furthermore, approximated trajectory costs in MDP-EI may exhibit considerable variance, even when the budget for exogenous scenarios is high, which could result in suboptimal actions being identified as simulation-based actions (notably, some suboptimal actions may have costs very close to the optimal one). In such a context, enhancing the accuracy of the simulation-based action $\hat{\pi}^+(\s)$—to either match the optimal action $\pi^+(\s)$ or perform similarly—becomes challenging. To achieve this, DCL utilizes a variance control mechanism (CRN), which serves to mitigate the variance in trajectory costs. Additionally, DCL employs a state-of-the-art bandit algorithm\footnote{A naive approach, known as the \emph{uniform allocation} strategy, would distribute the same number of independent exogenous scenarios ($M$) to each action \citep[e.g.,][]{lagoudakis2003reinforcement, Lazaric}.} (SH) to allocate exogenous scenarios to actions that demonstrate greater potential for optimality, avoiding the waste of exogenous scenarios on the suboptimal actions. Algorithm \ref{alg:seqhalving} explains the specifics of SH and the way CRN is integrated within its algorithm. First, we analyze the effectiveness of CRN in mitigating variance in trajectory costs, followed by an explanation of the SH algorithm.

\begin{algorithm}
\caption{Simulator}\label{alg:seqhalving}
\begin{algorithmic}[1]
\State \textbf{Input}: $\mathcal{M}$, $\s$, $\pi$, $M$, $H$
\State \textbf{Initialize}: Total rollout budget: $B_\s = M |\A_{\s}|$, number of rollouts allocated to competing actions: $T = 0$, sum of the approximate trajectory costs for action $a$, $a \in \A_{\s}$: $\Tilde{Q}_\pi(\s, a) = 0$, initial set of competing actions: $\A_0 = \A_{\s}$
\For{$r=0, 1, \dots, \ceil*{\log_{2}{|\A_{\s}|}} - 1 $} \algorithmiccomment{Sequential Halving}
\State $t_r = \ceil*{\frac{B_\s}{|\A_r| \ceil*{\log_{2}|\A_{\s}|}}} $, the number exogenous scenarios assigned to each action in $\A_r$ 
\For{$t=1, \dots, t_r $}
\State Generate an exogenous scenario $\boldsymbol\xi$ by \eqref{eq:exoscenario}, $|\boldsymbol\xi| = H$  \algorithmiccomment{Common Random Numbers}
\ForAll{$ a \in \A_r$}
\State Find $\hat{Q}_{\pi}(\s,a|\boldsymbol\xi)$ by \eqref{eq:qvalestimate} 
\State $\Tilde{Q}_{\pi}(\s,a) = \Tilde{Q}_{\pi}(\s,a) + \hat{Q}_{\pi}(\s,a|\boldsymbol\xi)$ \label{ag2:qsum}
\EndFor
\EndFor
\State $T = T + t_r$ \label{ag2:rollouts}
\State $\hat{q}_{\pi}(\s,a) =\Tilde{Q}_{\pi}(\s,a) / T$, $\forall a \in \A_r$ 
\State $\A_{r+1} = \arg \min_{a \in \A_r}^{1..\ceil*{\A_r / 2}} \hat{q}_{\pi}(\s,a) $ \algorithmiccomment{eliminate half of the actions}
\EndFor
\State \textbf{Output}: action in $\A_{\ceil*{log_{2}{|\A_\s|}}}$ %
\end{algorithmic}
\end{algorithm}

\emph{CRN} is a variance reduction technique frequently used in simulation studies. This method employs the same sequence of random numbers across distinct scenarios to decrease variance and enhance the reliability of comparative evaluations \citep{law2000simulation}. In an inventory management application, this may entail comparing the approximated trajectory costs of different actions using the \emph{same} demand sequence rather than generating them independently. To understand this, consider that estimating $\argmin_{a\in\A_\s}\hat{q}_{\pi}(\s,a)$ involves evaluating whether $\hat{q}_\pi(\s,a)-\hat{q}_\pi(\s,a')$ is greater or less than zero for all pairs of actions $a,a'\in \A_\s$. To illustrate this, let us consider two actions $a,a'\in \A_\s$. Let $X$ denote the resulting estimator of $\hat{q}_\pi(\s,a)-\hat{q}_\pi(\s,a')$ for a given set of independent exogenous scenarios $\boldsymbol{\xi}_r$, for $r=1,\ldots,M$. In other words, $X=\frac{1}{M}\sum_{r=1}^{M}\big(\hat{Q}_\pi(\s,a|\boldsymbol{\xi}_r)-\hat{Q}_\pi(\s,a'|\boldsymbol{\xi}_r)\big)$. Hence, $\E[X]=\hat{q}_\pi(\s,a)-\hat{q}_\pi(\s,a')$ and
\begin{align}
\var[X] =\var[\hat{Q}_\pi(\s,a|\boldsymbol{\xi})]+\var[\hat{Q}_\pi(\s,a'|\boldsymbol{\xi})]-2\cov[\hat{Q}_\pi(\s,a|\boldsymbol{\xi}),\hat{Q}_\pi(\s,a'|\boldsymbol{\xi})]. \label{eq:variance}
 \end{align}

The covariance term, emerging due to the pairwise comparison of the approximate trajectory costs, can be substantial relative to the variance terms, particularly for an MDP-EI. As such, variance control has the potential to calibrate the variance of the estimator $X$ to an appropriate level for discerning whether action $a$ is superior to action $a'$. The potential benefits of CRN will be confirmed in numerical results in \S\ref{sec:Sensitivity}, and are illustrated by Example \ref{ex2:crn} in Appendix \ref{app:examples}. 

\emph{SH} is an algorithm for allocating resources under a fixed budget to identify the optimal action from a set. This problem has been extensively explored in the literature under the umbrella terms of \emph{best arm identification in multi-armed bandits} \citep[cf.][]{Bubeck2012} and \emph{ranking and selection under fixed budget} \citep[cf.][]{hong2021review}. Without having any hyperparameters for tuning, SH has emerged as a preferred method in this field, demonstrating its effectiveness in environments requiring fast resource allocation \citep[e.g.,][for its effect in reducing the simulation effort exhibited by AlphaZero, which previously utilized Upper Confidence Bounds method]{danihelka2022policy}.

In DCL, SH assigns exogenous scenarios to the more promising actions based on their empirical average expected costs, while concurrently accounting for the aforementioned variance control scheme (CRN). Specifically, given a state $\s$ and the feasible action set $\mathcal{A}_\s$, the algorithm evenly distributes the total budget $B_\s$ across $\log_{2}{|\A_\s|}$ rounds and eliminates the lower-performing half of the actions at the end of each round. Within a round, the actions are allocated an identical number of exogenous scenarios. For each scenario, we retain the exogenous inputs (CRN) and calculate the approximate trajectory costs \eqref{eq:qvalestimate} for the competing actions in that round. 

DCL diverges from the original SH algorithm \citep[see][]{seqhalving} in terms of statistics storage. Rather than discarding the collected approximate trajectory costs of the competing actions at the end of a round, DCL \emph{stockpiles} \citep{seqhalvingscores} them by maintaining the sum of the trajectory costs for the actions, such that $\Tilde{Q}_{\pi}(\s,a) = \sum_{i=1}^{T} \hat{Q}_{\pi}(\s,a|\boldsymbol\xi_i)$, where $T$ represents the number of exogenous scenarios assigned for action $a \in \A_r$ until the end of round $r$, and $\A_r$ denotes the set of competing actions in round $r$ (cf. Line \ref{ag2:qsum} in Algorithm \ref{alg:seqhalving}). This helps to decrease the variance term \eqref{eq:variance} further as we collect more trajectory costs for the actions.

After the last round of SH, the single remaining action is identified as the simulation-based action $\hat{\pi}^+(\s)$, corresponding to the sampled state $\s$. This state-action pair is subsequently incorporated into the dataset. DCL can efficiently construct its dataset $\K_i=\{(\s_k,\hat{\pi}_i^+(\s_k))|k\in \{1,\ldots,N\}\}$ at iteration $i$ by systematically following this process, which entails sampling a state and determining its simulation-based action. This dataset is then utilized in the subsequent stage of training neural networks to formulate a policy representation.

\noindent \textbf{Policy approximation with neural networks}

Upon construction of the dataset $\K_i=\{(\s_k,\hat{\pi}^+(\s_k))|k\in \{1,\ldots,N\}\}$, DCL trains a neural network to approximate the simulation-based policy $\hat{\pi}^+(\cdot)$. Specifically, given a predefined network structure $\NN_\theta$, the objective is to update the parameters $\theta$ such that the resulting policy $\pi_\theta(\cdot)$ closely aligns with the simulation-based policy $\hat{\pi}^+(\cdot)$ within the states present in $\K_i$. With a sufficiently large number of samples ($N$) and appropriate training, the neural network may provide generalization across the entire state space $\mathcal{S}$ \citep{boute2021deep}. Unlike other versions of approximate policy iteration, our machine learning model solves a classification task, which may contribute to stable training and good generalization. For training neural networks, we adopt a standard set of strategies \citep[see][]{goodfellow2016deep}; we refer readers to Appendix \ref{app:neuralnetworktraining} for the details, including the specifics of the $Classifier$ algorithm (Line \ref{alg1:classifier} in Algorithm \ref{alg:DCL}).

After completion of training, the neural network policy is defined as $\pi_\theta(\s)=\argmax_{a\in \A_{\s}} (\NN_{\theta}(\s)[a])$, valid for $\forall \s \in \mathcal{S}$, and this policy is the initial policy in the next approximate policy iteration step, $i+1$, continuing until $\pi_n$ is obtained.

\section{Experimental Setup}\label{sec:experimentsetup}

We next present the experimental setup employed to apply DCL to three difficult inventory problems: lost sales inventory control, perishable inventory systems, and inventory systems with random lead times. While we discuss important characteristics of the inventory settings and the features of the problem instances in \S\ref{sec:results}, we discuss modeling them under MDP-EI, $\mathcal{M} = \langle \mathcal{S}, \A, \mathcal{W}, \Input, f, C, \alpha, \s_{0} \rangle$, in Appendix~\ref{app:invmodelformulation}. For each system, a BSP is adopted as $\pi_0$, see also the appendix. 

\begin{table}[ht]
  \centering
  \footnotesize
  \begin{tabular}{|ll|ll|}
\midrule
\multicolumn{2}{|c|}{Sampling and Simulation} & \multicolumn{2}{c|}{Neural Network Structure $\NN_\theta$} \\
\midrule
Horizon length:       &$H = 40$   & Number of layers:  & 4\\ 
Number of exogenous scenarios:    &$M = 1000$ & Number of neurons: & \{ 256, 128, 128, 128 \}  \\
Number of samples:   &$N = 5000$ & Optimizer:       & Adam \\
Length of the warm-up period:       &$L = 100$   &Mini-batch size:  & $MiniBatchSize = 64$\\
\midrule
\multicolumn{4}{|c|}{Number of approximate policy iterations: $n = 3$} \\
    \bottomrule
    \end{tabular}
\caption{The hyperparameters used in the experiments.}
\label{tab:hyperparameters}
\end{table}

The DCL algorithm is implemented in C++20; computations are performed on an AMD EPYC 7H12 processor with 128 hardware threads. Table \ref{tab:hyperparameters} outlines the hyperparameters employed for all numerical experiments. We maintain consistency of the hyperparameters throughout the experiments to demonstrate the robustness of DCL across different models and their respective problem instances. Since exact policy iteration usually converges to near-optimal solutions within 3-6 steps for stochastic models \citep{puterman2014markov}, we opt for $n=3$ approximate policy iteration steps to ensure computational efficiency while still maintaining a high likelihood of approaching optimality. The sampling-related hyperparameters, $N$ and $L$, are chosen to align with the discussion in \S\ref{sec:dcl}. We select sufficiently high values for simulation parameters $H$ and $M$ to ensure accurate estimates of the action-value functions. We use a multi-layer perceptron as neural network architecture and use standard values for training hyperparameters, see Appendix \ref{app:neuralnetworktraining} for details.

For reproducibility purposes, we share the source code and the trained weights of neural networks for each experiments on our github page \href{https://github.com/tarkantemizoz/DynaPlex}{https://github.com/tarkantemizoz/DynaPlex}.

\section{Results}\label{sec:results}

We compare the performance of DCL against BSP and the leading heuristics in three inventory management problems: lost sales inventory control, perishable inventory systems and inventory system with random lead times. During our preliminary assessments, we consider the PPO algorithm as a potential benchmark. However, despite tuning its hyperparameters, PPO does not demonstrate clear superiority over the best heuristics, let alone BSP. This leads us to rule out PPO as a benchmark for our experiments. Our quantitative analysis then focuses on how the inclusion of SH and CRN affect the performance of DCL, aiming to provide insights into its efficiency.

\subsection{Performance Evaluation on Inventory Problems}\label{sec:mainresults}

For our numerical experiments, we use the problem instances previously utilized to measure the efficacy of the best heuristics in their original papers. We refer readers to those benchmark papers for the theoretical and practical relevance of the selected instances. These instances include both situations where the computation of the optimal policy is feasible and more complex cases where the real benefit of approximation methods, including DCL, becomes evident. For the remainder of our results, the instances will be referred to as "small instances" and "large instances", respectively. 

Depending on how the results are presented in the benchmark papers, we report optimality gaps (only for small instances), relative cost improvements over BSP and average costs per period. Optimality gaps are computed as $(v_\pi-v^*)/v^*\times 100\%$, where $v_\pi$ denotes the average cost per time unit of the policy $\pi$, and $v^*$ stands for the average cost per time unit of the optimal policy. Here, $v_\pi$ and $v^*$ are computed by explicitly solving the average-cost Bellman equations \citep{puterman2014markov}. The relative cost improvement of DCL over BSP can be calculated by the formula $(\hat{v}_{\pi_{DCL}}-\hat{v}_{\pi_{BSP}})/\hat{v}_{\pi_{BSP}}\times 100\%$, with $\hat{v}_{\pi_{DCL}}$ denoting average cost per period of DCL and $\hat{v}_{\pi_{BSP}}$ denoting average cost per period of BSP, respectively. We obtain unbiased estimators of the average costs per period through simulations. Each evaluation is comprised of 1000 runs, with each run spanning 5000 periods and starting after following a warm-up period of 100 periods. Results are statistically significant: the half-width of a 95\% confidence interval is less than 1\% of the corresponding cost value.

DCL generates three neural network policies by implementing three approximate policy iteration steps ($n=3$). We report the performance of the best-performing neural network for each instance, which is typically (but not invariably) the third/last generation. 

\noindent \textbf{Lost sales inventory control}

We consider the well-known discrete-time lost sales inventory system \citep[][]{zipkin2008old}. In this system, we manage an inventory of items over time by placing orders for new items, attempting to meet the exogenous demands, and incurring costs based on whether we have sufficient inventory to meet the demand or not (in case of lost sales, partial fills are assumed). The constant lead time for new orders to arrive is greater than one period (i.e., $\tau>1$). Besides BSP, the benchmarks include the best-performing capped base-stock (CBS) policy and myopic-2-period (M-2) policy, which provides better optimality gaps in some cases. We also report the results from a previous DRL application to such systems, the A3C algorithm by \citet{gijsbrechts2019can}.

We employ instances from the testbed proposed by \citet{zipkin2008old} and \citet{Xin2021}. Specifically, we maintain a constant holding cost of $h=1$ and adjust the penalty cost $p$ to encompass the range $p = \{4,9,19,39\}$. We apply two types of demand distributions, Poisson and geometric, both with a mean of five. For small instances, we set the lead time $\tau$ to $\{2,3,4\}$, while for large instances, we adopt lead times of $\tau =\{6,8,10\}$.

\DeclareFontSeriesDefault[rm]{bf}{b}

\begin{table}[ht]
  \centering
  \footnotesize

    \begin{tabular}{|r|l|ccc|ccc|ccc|ccc|}
\cmidrule{3-14}    \multicolumn{2}{c|}{\multirow{2}[4]{*}{}} & \multicolumn{6}{c|}{Poisson} & \multicolumn{6}{|c|}{Geometric} \\
\cmidrule{3-14}    \multicolumn{2}{c|}{} & \multicolumn{6}{c|}{Lead time $\tau$} & \multicolumn{6}{|c|}{Lead time $\tau$} \\
    \midrule
    \multicolumn{1}{|r|}{p} & Policy  &  2     & 3     & 4 & 6 & 8 & \multicolumn{1}{c|}{10}     & \multicolumn{1}{|c}{2}     & 3     & 4  & 6 & 8 & 10  \\
    \midrule
    \multirow{5}[2]{*}{$4$} & BSP &
    \multicolumn{1}{r}{5.5\%} & 
    \multicolumn{1}{r}{8.2\%} & 
    \multicolumn{1}{r|}{9.9\%} & 5.51 & 5.72 & 5.86 &
    \multicolumn{1}{|r}{4.5\%} &
    \multicolumn{1}{r}{6.4\%} & 
    \multicolumn{1}{r|}{7.8\%} & 11.86 & 12.12 & 12.31 \\
        & CBS   & 
    \multicolumn{1}{r}{0.2\%} & 
    \multicolumn{1}{r}{0.7\%} & 
    \multicolumn{1}{r|}{1.5\%} &  5.03 & 5.19 & 5.27 &
    \multicolumn{1}{|r}{0.8\%} & 
    \multicolumn{1}{r}{0.4\%} & 
    \multicolumn{1}{r|}{0.8\%} & 10.91 & 10.96 & 10.98\\
      & M-2   & 
    \multicolumn{1}{r}{0.2\%} & 
    \multicolumn{1}{r}{0.8\%} & 
    \multicolumn{1}{r|}{1.9\%} &  5.05 & 5.20 & 5.31 &
    \multicolumn{1}{|r}{0.5\%} & 
    \multicolumn{1}{r}{1.2\%} & 
    \multicolumn{1}{r|}{1.7\%} & 11.08 & 11.27 & 11.40  \\
         & A3C &
    \multicolumn{1}{r}{3.2\%} &
    \multicolumn{1}{r}{3.0\%} &
    \multicolumn{1}{r|}{6.7\%}&$-$&$-$&$-$ &
    \multicolumn{1}{|r}{$-$} &
    \multicolumn{1}{r}{$-$} &
    \multicolumn{1}{r|}{$-$}&$-$&$-$&$-$ \\
      & DCL &
    \multicolumn{1}{r}{0.01\%} &
    \multicolumn{1}{r}{\textbf{0.01\%} } &
    \multicolumn{1}{r|}{0.03\%} & 4.88 & 4.96 & \textbf{5.02} &
    \multicolumn{1}{r}{\textbf{0.01\%}} &
    \multicolumn{1}{r}{\textbf{0.01\%} } &
    \multicolumn{1}{r|}{\textbf{0.02\%}} & \textbf{10.75} & 10.86 & \textbf{10.87} \\
          & DCL-20000 &
    \multicolumn{1}{r}{\textbf{0.00\%}} &
    \multicolumn{1}{r}{\textbf{0.01\%} } &
    \multicolumn{1}{r|}{\textbf{0.01\%}} & \textbf{4.87} & \textbf{4.95} & \textbf{5.02} &
    \multicolumn{1}{|r}{\textbf{0.01\%}} &
    \multicolumn{1}{r}{\textbf{0.01\%}} &
    \multicolumn{1}{r|}{\textbf{0.02\%}} & \textbf{10.75} & \textbf{10.84} & \textbf{10.87} \\
        \midrule
    \multirow{5}[2]{*}{$9$} & BSP &
    \multicolumn{1}{r}{3.7\%} &
    \multicolumn{1}{r}{5.1\%} &
    \multicolumn{1}{r|}{6.4\%} & 7.90 & 8.32 & 8.63 &
    \multicolumn{1}{|r}{3.1\%} &
    \multicolumn{1}{r}{4.6\%} &
    \multicolumn{1}{r|}{5.8\%} & 18.53 & 19.18 & 19.68\\
            & CBS   & 
    \multicolumn{1}{r}{0.5\%} & 
    \multicolumn{1}{r}{1.4\%} & 
    \multicolumn{1}{r|}{1.0\%} & 7.26 & 7.55 & 7.77 &
    \multicolumn{1}{|r}{0.8\%} & 
    \multicolumn{1}{r}{0.8\%} & 
    \multicolumn{1}{r|}{0.9\%} & 17.35 & 17.68 & 17.88\\
      & M-2   &
    \multicolumn{1}{r}{0.2\%} &
    \multicolumn{1}{r}{0.6\%} &
    \multicolumn{1}{r|}{1.2\%} &  7.43 & 7.77 & 8.08 &
    \multicolumn{1}{|r}{0.5\%} &
    \multicolumn{1}{r}{1.1\%} &
    \multicolumn{1}{r|}{1.9\%}  & 17.75 & 18.39 & 18.89\\
         & A3C & 
    \multicolumn{1}{r}{4.8\%} &
    \multicolumn{1}{r}{3.1\%} &
    \multicolumn{1}{r|}{3.4\%} &$-$&$-$&$-$&
    \multicolumn{1}{|r}{$-$} &
    \multicolumn{1}{r}{$-$} & 
    \multicolumn{1}{r|}{$-$} &$-$&$-$&$-$\\
      & DCL &
    \multicolumn{1}{r}{\textbf{0.00\%}} &
    \multicolumn{1}{r}{0.03\%} &
    \multicolumn{1}{r|}{0.06\%} & \textbf{7.23} & 7.52 & 7.69 &
    \multicolumn{1}{r}{\textbf{0.00\%}} &
    \multicolumn{1}{r}{\textbf{0.01\%} } &
    \multicolumn{1}{r|}{0.04\%} & \textbf{17.13} & \textbf{17.48} & \textbf{17.71} \\
              & DCL-20000 &
    \multicolumn{1}{r}{\textbf{0.00\%}} &
    \multicolumn{1}{r}{\textbf{0.01\%}}  &
    \multicolumn{1}{r|}{\textbf{0.01\%}} & \textbf{7.23} & \textbf{7.49} & \textbf{7.68} &
    \multicolumn{1}{|r}{0.01\%} &
    \multicolumn{1}{r}{\textbf{0.01\%}} &
    \multicolumn{1}{r|}{\textbf{0.01\%}} & \textbf{17.13} & \textbf{17.48} & \textbf{17.71} \\
        \midrule
    \multirow{4}[2]{*}{$19$} & BSP & 
    \multicolumn{1}{r}{2.3\%} &
    \multicolumn{1}{r}{2.9\%} &
    \multicolumn{1}{r|}{3.9\%} & 10.20 & 10.90 & 11.48 &  
    \multicolumn{1}{|r}{2.0\%} &
    \multicolumn{1}{r}{3.0\%} &
    \multicolumn{1}{r|}{3.9\%} & 25.54 & 26.81 & 27.82\\
            & CBS   & 
    \multicolumn{1}{r}{0.8\%} & 
    \multicolumn{1}{r}{0.5\%} & 
    \multicolumn{1}{r|}{0.7\%} & 9.80 & 10.35 & 10.66 &  
    \multicolumn{1}{|r}{0.8\%} & 
    \multicolumn{1}{r}{1.0\%} & 
    \multicolumn{1}{r|}{1.4\%}  & 24.49 & 25.38 & 25.98\\
      & M-2   &
    \multicolumn{1}{r}{0.1\%} &
    \multicolumn{1}{r}{0.4\%} &
    \multicolumn{1}{r|}{0.8\%}  & 9.85 & 10.53 & 11.09 & 
    \multicolumn{1}{|r}{0.3\%} &
    \multicolumn{1}{r}{0.8\%} &
    \multicolumn{1}{r|}{1.4\%}  & 24.93 & 26.21 & 27.27  \\
      & DCL &
    \multicolumn{1}{r}{0.01\%} &
    \multicolumn{1}{r}{0.03\% } &
    \multicolumn{1}{r|}{0.06\%} &  9.66 & 10.20 & 10.76 &
    \multicolumn{1}{r}{\textbf{0.01\%}} &
    \multicolumn{1}{r}{0.02\%}  &
    \multicolumn{1}{r|}{0.04\%} & 24.21 & 25.11 & 26.14 \\
                  & DCL-20000 &
    \multicolumn{1}{r}{\textbf{0.00\%}} &
    \multicolumn{1}{r}{\textbf{0.01\%} } &
    \multicolumn{1}{r|}{\textbf{0.01\%}}  &  \textbf{9.65} & \textbf{10.18} & \textbf{10.57} &
    \multicolumn{1}{|r}{\textbf{0.01\%}} &
    \multicolumn{1}{r}{\textbf{0.01\%}} &
    \multicolumn{1}{r|}{\textbf{0.01\%} }& \textbf{24.20} & \textbf{25.07} & \textbf{25.71} \\
        \midrule
    \multirow{4}[2]{*}{$39$} &
    BSP &
    \multicolumn{1}{r}{0.9\%} &
    \multicolumn{1}{r}{1.8\%} &
    \multicolumn{1}{r|}{2.5\%} &  12.38 & 13.39 & 14.24 &
    \multicolumn{1}{|r}{1.3\%} &
    \multicolumn{1}{r}{2.0\%} & 
    \multicolumn{1}{r|}{2.6\%}  & 32.69 & 34.47 & 36.25\\
            & CBS   & 
    \multicolumn{1}{r}{0.3\%} & 
    \multicolumn{1}{r}{0.4\%} & 
    \multicolumn{1}{r|}{0.8\%} &  12.08 & 12.94 & 13.71 &
    \multicolumn{1}{|r}{0.3\%} & 
    \multicolumn{1}{r}{1.1\%} & 
    \multicolumn{1}{r|}{1.4\%}  & 31.86 & 33.97 & 35.64 \\
      & M-2   &
    \multicolumn{1}{r}{0.1\%} &
    \multicolumn{1}{r}{0.3\%} &
    \multicolumn{1}{r|}{0.4\%} & 12.11 & 13.09 & 13.93 &
    \multicolumn{1}{|r}{0.2\%} & 
    \multicolumn{1}{r}{0.5\%} &
    \multicolumn{1}{r|}{0.9\%} & 32.12 & 34.12 & 35.82 \\
       & DCL &
    \multicolumn{1}{r}{0.01\%} &
    \multicolumn{1}{r}{0.02\%} &
    \multicolumn{1}{r|}{0.09\%} & 11.96 & 12.83 & 13.96 &
    \multicolumn{1}{|r}{0.02\%} &
    \multicolumn{1}{r}{0.03\%} &
    \multicolumn{1}{r|}{0.04\%} & 31.47 & 33.10 & 35.59\\
                  & DCL-20000 &
    \multicolumn{1}{r}{\textbf{0.00\%}} &
    \multicolumn{1}{r}{\textbf{0.01\%}}  &
    \multicolumn{1}{r|}{\textbf{0.02\%}} & \textbf{11.95} & \textbf{12.81} & \textbf{13.49} &
    \multicolumn{1}{|r}{\textbf{0.01\%}} &
    \multicolumn{1}{r}{\textbf{0.01\%} }&
    \multicolumn{1}{r|}{\textbf{0.02\%}} &  \textbf{31.46} & \textbf{33.05} & \textbf{34.29} \\
    \bottomrule
    \end{tabular}%
       \caption{Optimality gaps (for $\tau=\{2,3,4\}$) and average costs per period (for $\tau=\{6,8,10\}$) of selected policies on lost sales inventory control, with the best-performing policies for each problem \textbf{highlighted} bold. Results for A3C are reported where exact numerical results are available in the respective paper \citep[see][]{gijsbrechts2019can}.}
  \label{tab:lostsalesresults}%
\end{table}%

Table \ref{tab:lostsalesresults} illustrates the optimality gaps for small instances and the average costs incurred by the policies for large instances. We report the optimality gaps of the benchmarks to one decimal place. The gaps for DCL are reported to two decimal places to retain a sense of the approximate magnitude of the optimality gap. For small cases, the table shows that DCL always achieves an optimality gap less than $0.09\%$, indicating that the learned neural network successfully encapsulates the structure of the optimal policy to such an extent that it nearly matches its performance. Moreover, DCL outperforms other methods, including A3C algorithm, by achieving an optimality gap at least ten times smaller. In many instances, we also observe a small increase in the optimality gaps for DCL as the lead time grows.

With regard to the average costs per period for larger cases, the table shows that DCL yields lower costs than other methods for each instance, except for cases where $\tau=10$, $p\in \{19,39\}$ with Poisson distributed demands and where $\tau=10$, $p=19$ with geometric distributed demands. We hypothesize that the number of samples ($N=5000$) may be insufficient to cover the state space comprehensively for these instances. Therefore, we also train DCL with $20000$ samples instead of $5000$ while keeping other hyperparameters fixed and denote this model as DCL-20000. DCL-20000 achieves a fair reduction in costs in the three instances where the original DCL could not outperform the CBS policy, thereby surpassing all heuristic results. On the other hand, the improvement of DCL-20000 over DCL seems insignificant in many cases, showing that DCL can already obtain good results (close to optimal) despite trained with less number of samples. 

Computation times to apply DCL for individual instances vary between $100$ and $250$ seconds, while DCL-20000 takes between $7$ and $17$ minutes. This demonstrates a significant improvement when contrasted with applying the A3C algorithm to inventory problems, which took days due to the need for extensive hyperparameter tuning. (When assessing this computation time and comparing it to computation times reported by others, one should consider the capabilities of different computing machines regarding the number of threads and their computational power and the possible availability of accelerators (not available in our case).) We also find that the training of neural networks is consistently completed within $15$ and $40$ seconds for DCL and DCL-20000, respectively. The remainder of the computation time can be largely attributed to the simulations. These are observed to consume more time when the lead time was longer and when the demand followed a geometric distribution. This is to be expected, as these conditions require evaluating a larger number of actions, leading to creating a greater number of exogenous scenarios because the total budget is $B_\s=M|\A_\s|$ for a state $\s$.

Considering all these results, it becomes clear that DCL sets new performance benchmarks for the well-established testbed of the canonical lost sales inventory control problem.

\noindent \textbf{Perishable inventory systems}

We next turn our attention to the performance of DCL on perishable inventory systems. In perishable inventory systems, a product possesses a maximum life of $m_l$ periods and having a constant lead time $\tau \geq 0$. The unmet demands are lost and penalized, and the products over their life time are scrapped and result in incurring waste costs. The benchmark policies for comparison include BSP and the BSP-low-EW policy by \citet{Haijema2019}.

In order to evaluate the performances of the policies, we have selected a subset comprising 81 diverse instances from the extensive testbed of 11177 instances utilized by \citet{Haijema2019}. Our selection focuses on those settings that require non-trivial policies, ensuring a robust and comprehensive assessment. The holding cost is set at $h=0$, the penalty cost at $p=100$, and the waste cost at $w=100$. We maintain a constant total mean demand per period of $\mu=4$, while the coefficient of variation $cvr$ ranges between $\{1.0,1.5,2.0\}$. On average, a fraction $f$ of demand is fulfilled by the FIFO issuing policy, with the remaining fraction $1-f$ satisfied by the LIFO policy. This is modeled through two discrete demand distributions, fitted using the method proposed by \citet{adan1995}. (With the utilized mean and coefficient of variation levels, poisson, negative binomial and geometric distribution can be fitted.) In the experiments, the probability $f$ varies between $f = \{1.0,0.5,0.0\}$. We also set the product lifetime $m_l=\{3,4,5\}$ and order lead time $\tau=\{0,1,2\}$. Small instances correspond to cases where $m_l+\tau \leq 5$ for $f=0$ and $f=1$, and $m_l+\tau < 4$ for $f=0.5$. The difference arises from the fact that $f=0.5$ requires two fitted demand distributions, which increases the possible state transitions.  Large instances correspond to the remaining cases. 

\begin{table}[ht]
  \centering
  \footnotesize
\begin{tabular}{|c|c|cc|ccc|cc|}
\cmidrule{2-9}
\multicolumn{1}{c}{\multirow{1}[4]{*}{}} & \multicolumn{3}{|c|}{All Instances}  & \multicolumn{3}{c|}{Small Instances} & \multicolumn{2}{c|}{Large Instances}\\
\cmidrule{2-9}
\multicolumn{1}{c}{\multirow{1}[4]{*}{}} & \multicolumn{1}{|c|}{Avg. Cost}& \multicolumn{2}{c|}{Avg. BSP gap}  & \multicolumn{3}{c|}{Avg. Opt. gap} &\multicolumn{2}{c|}{Avg. BSP gap}\\
 \midrule
    \multicolumn{1}{|l|}{Instances} & BSP& BSP-low-EW & DCL & BSP & BSP-low-EW & DCL & BSP-low-EW & DCL \\
    \midrule
$All$       & 75.3& -4.6\% & -8.6\% & 9.7\%  & 4.1\% & 0.03\%  & -4.3\% & -8.9\% \\
 \midrule
$m_l = 3$     & 99.7 & -6.5\%& -8.7\% & 9.1\% & 2.6\%  & 0.02\%  & -11.7\%& -13.1\%  \\
$m_l = 4$     & 71.7& -4.4\% & -8.5\%  & 9.2\% & 5.3\% & 0.04\% & -5.6\% & -8.9\%  \\
$m_l = 5$     & 54.3& -3.0\% & -8.7\% & 13.0\% & 7.5\% & 0.04\% & -2.6\% & -8.2\%  \\
$\tau = 0$     & 57.4 & -3.2\% & -7.6\% & 9.1\%  & 4.9\% & 0.02\%  & -0.5\% & -6.1\% \\
$\tau = 1$     & 77.5& -3.9\%& -7.7\%  & 8.8\%  & 3.9\% & 0.03\%  & -3.2\%& -7.2\% \\
$\tau = 2$     & 90.8& -6.8\%  & -10.6\% & 13.9\% & 1.6\% & 0.06\% & -5.6\%  & -10.2\%\\
$cvr = 1.0$ & 37.2& -4.5\% & -9.3\% & 10.9\%  & 5.0\% & 0.03\%  & -3.8\% & -9.3\%  \\
$cvr = 1.5$ & 74.0& -4.9\% & -8.8\%  & 9.9\%  & 4.3\% & 0.05\%  & -4.8\% & -9.0\% \\
$cvr = 2.0$ & 114.5& -4.5\%   & -7.8\% & 8.2\%  & 3.1\% & 0.02\%  & -4.4\%   & -8.3\%\\
$f = 1.0$   & 47.9 & -3.7\%  & -6.7\% & 6.3\%  & 2.8\% & 0.02\% & -4.6\%  & -8.6\% \\
$f = 0.5$   & 74.1 & -3.7\% & -6.8\%& 5.1\% & 3.0\%  & 0.04\% & -4.5\% & -7.8\% \\
$f = 0.0$   & 103.7 & -6.6\%  & -12.4\%& 15.3\%  & 6.0\%  & 0.04\% & -3.7\%  & -11.2\%\\
\bottomrule
\end{tabular}
\caption{Average costs of BSP, the optimality gaps, and the difference in average costs between BSP and DCL - BSP-low-EW.}
\label{tab:perishableresults}
\end{table}

In line with the results in \citet{Haijema2019}, we average the results according to identical experimental parameters. Table \ref{tab:perishableresults} provides an overview of these findings, reporting the average cost of BSP, the optimality gaps of the three policies for small instances, and the relative cost improvement of DCL and BSP-low-EW policies over BSP for large instances and also for all instances. We observe that DCL substantially outperforms both policies both for small and large instances. For small instances, it achieves a markedly lower average optimality gap ($0.03\%$) and consistently delivers robust performance across different settings, including various issuing policies. (When analyzing instances individually, we observe at most $0.2\%$ optimality gap for DCL.) For large instances, DCL performs considerably better than BSP as well as BSP-low-EW. We further note that the relative cost improvement of DCL over BSP tends to increase with longer lead times and with an increase in demand met through the LIFO issuing policy when averaged over all instances. 

Similar to our lost sales inventory control experiments, the computation times to apply DCL for individual instances range between $150-200$ seconds and tend to increase as $m_l+\tau$ expands.

\noindent \textbf{Inventory systems with random lead times}

Finally, we evaluate the performance of DCL on inventory systems with random lead times. We consider a continuous review inventory system with independent, identically distributed replenishment lead times with order-crossing and backlogged unsatisfied demands. We assume demands follow poisson process with mean rate $\mu$. Our benchmark policies are BSP and the generalized base-stock (GBS) policy \citep{Stolyar2022}.

We consider the instances adopted by \citet{Stolyar2022} for numerical experiments, which comprise cases with three distinct lead time distributions: exponential (small instances), pareto and uniform (large instances). For all lead time distribution cases, we first maintain the holding cost $h$, and the backorder cost $b$ at a constant value of 1, and vary the ratio $\mu / \beta$ which takes values in $\{2, 10, 20\}$, where $\beta$ denotes the mean lead time. Then, we replicate the sensitivity analysis performed by \citet{Stolyar2022} for the exponential distribution case. We keep $\mu / \beta =20$ constant and alternate one of the holding cost $h$, or the backorder cost $b$ within the range $\{3,6,9\}$ while maintaining the other at 1. Lastly, we keep $h=1$ and $\mu / \beta =20$, and vary $b$ within the range $\{9,19,39,69,99\}$ for uniform and pareto distibution cases to reflect more realistic cases. We note that \citet{Stolyar2022} do not consider these cases. We set the maximum order quantity at a decision epoch to $m=6$ as GBS policy is likely to order less than this number in the selected problem instances. 

\begin{table}[ht]
  \centering
  \footnotesize
  \begin{tabular}{|r|r|ccc|cccccc|}
\cmidrule{2-11}
\multicolumn{1}{c|}{\multirow{1}[3]{*}{}} & $\mu / \beta$      & 2 & 10 & 20  & 20 & 20   & 20  & 20   & 20   &  20 \\
\multicolumn{1}{c|}{\multirow{1}[3]{*}{}}&$h$ & 1 & 1  & 1   &  9 &  6   & 3   &  1   & 1    & 1 \\
\multicolumn{1}{c|}{\multirow{1}[3]{*}{}}&$b$  & 1 & 1  & 1   &  1 &  1   & 1   &  3   & 6    & 9  \\
\midrule
\multirow{3}[2]{*}{Exponential lead time} & BSP & 1.08 & 2.50 & 3.55 & 7.45 & 6.75 & 5.52 & 5.79 & 7.32 & 8.17 \\
&DCL & 0.95 & 1.87 & 2.46 & 5.47 & 4.91 & 3.94 & 3.92 & 4.89 & 5.42 \\
&GBS & 1.00 & 2.01 & 2.66 & 5.62 & 5.22 & 4.17 & 4.18 & 5.14 & 5.58 \\
&Optimal & 0.95 & 1.87 & 2.45 & & & & & & \\
\midrule
\multicolumn{1}{c|}{\multirow{1}[3]{*}{}} & $\mu / \beta$      & 2 & 10 & 20  & 20 & 20   & 20  &  20  & 20   &   \\
\multicolumn{1}{c|}{\multirow{1}[3]{*}{}}&$h$ & 1 & 1  & 1   &  1 &  1   & 1   &  1   & 1    &  \\
\multicolumn{1}{c|}{\multirow{1}[3]{*}{}}&$b$  & 1 & 1  & 1   &  9 &  19   & 39   &  69   & 99    &  \\
\midrule
\multirow{3}[2]{*}{Uniform lead time}& BSP & 1.08 & 2.50 & 3.56   &8.17 &9.80 &11.15 &12.25 &12.86 &\\
&DCL & 1.02 & 2.15 & 2.90   &6.74 &8.05 &9.41 &10.61 &11.31 &\\
&GBS & 1.06 & 2.29 & 3.13   & & & & & &\\
\midrule
\multirow{3}[2]{*}{Pareto lead time}& BSP & 1.07 & 2.49 & 3.56   &8.18 &9.73 &11.16 &12.28 &12.88 &\\
&DCL & 0.87 & 1.71 & 2.37   &5.62 &5.86 &7.07 & 7.34 & 8.01 &\\
&GBS & 0.96 & 1.93 & 2.47   & & & & & &\\
    \bottomrule
    \end{tabular}%
\caption{Average costs of policies when lead times are exponentially, uniformly or pareto distributed.}
\label{tab:rlt}%
\end{table}%

Table \ref{tab:rlt} reports the average costs per period of the policies across the selected instances, showcasing the dominant performance of DCL. In cases of exponential lead time, we observe that DCL nearly mirrors the average cost per period of the optimal policy, suggesting the capability of DCL in grasping the structure of the optimal policy under such settings. For instances with uniform and pareto lead times, DCL retains its superiority over GBS. We observe that the difference in average costs per period between BSP and DCL narrows for uniform lead times and grows for pareto lead times. The gap between BSP and GBS likewise contracts. This contraction might be ascribed to the inherent nature of the uniform distribution, which has an increasing hazard rate, potentially negatively affecting the development of proficient policies \citep{Stolyar2022}. Lastly, DCL significantly outperforms BSP with varying $b/h$ ratios for all distributions.

The computation times for applying DCL to individual instances vary between $150-300$ seconds. 

\subsection{Impact of SH with CRN on Performance Improvements}\label{sec:Sensitivity}

The numerical results reveal that DCL is a powerful framework for addressing intractable inventory management problems, outdoing the top-performing heuristics. Next, we focus on quantifying the benefits of incorporating SH with CRN in the simulation phase of the algorithm. To investigate the impact of these techniques on the efficiency of DCL, we rerun the numerical experiments for the three inventory problems. This time, however, we employ DCL without incorporating them. Specifically, in the process of determining the simulation-based action $\hat{\pi}^+(\s)$ for any given state $\s$, DCL distributes the total exogenous scenario budget uniformly among each applicable action $a \in \A_\s$ – a methodology referred to as uniform allocation – and simultaneously alternates the scenarios among the actions, which implies the absence of CRN. This variant of the DCL algorithm is referred to as DCL$_0$.

Table \ref{tab:Sensitivity} provides a comprehensive overview of the optimality gaps (for small instances) and the relative cost improvements of the policies over BSP regarding average costs (BSP gap for large instances and for all instances). These values represent averages across all respective instances for each inventory setting \footnote{For inventory systems with random lead times, we only consider instances where the results for the GBS policy is available.}, focusing on DCL, DCL$_0$, and $\pi_{heur}$, the latter being the best heuristic for each corresponding inventory problem.

\begin{table}[ht]
  \centering
  \footnotesize
  \begin{tabular}{|r|rrr|rrr|rrr|}
\cmidrule{2-10}
\multicolumn{1}{c}{\multirow{1}[4]{*}{}} & \multicolumn{3}{|c|}{Lost Sales} & \multicolumn{3}{c|}{Perishable} &  \multicolumn{3}{c|}{Random Lead Times}  \\
\cmidrule{2-10}
\multicolumn{1}{c|}{\multirow{1}[3]{*}{}}&  \multicolumn{1}{c}{DCL} &  \multicolumn{1}{c}{DCL$_0$} & \multicolumn{1}{c|}{$\pi_{heur}$} &   \multicolumn{1}{c}{DCL} & \multicolumn{1}{c}{DCL$_0$} & \multicolumn{1}{c|}{$\pi_{heur}$} &   \multicolumn{1}{c}{DCL} & \multicolumn{1}{c}{DCL$_0$} & \multicolumn{1}{c|}{$\pi_{heur}$} \\
\midrule
Optimality gap  & 0.03\%& 1.3\%& 0.8\%& 0.03\%& 0.9\%& 4.1\%& 0.14\%& 0.8\%& 7.1\%\\
BSP gap (large instances) & -7.5\%& -0.2\%& -5.4\%& -8.9\%& -7.6\%&  -4.3\%& -20.3\%& -19.5\%& -14.3\%\\
BSP gap (all instances) & -5.7\%& -1.4\%& -4.8\%& -8.6\%& -7.6\%&  -4.6\%& -24.7\%& -23.9\%& -19.9\%\\
    \bottomrule
    \end{tabular}
\caption{Performance comparison of DCL, DCL$_0$ and the best heuristics. (The optimality gaps for Random Lead Times are approximated by comparing the average cost per period of DCL ($\hat{v}_{\pi_{DCL}}$), obtained through simulations, with the average cost per period of the optimal policy reported by \citet{Stolyar2022}.) }
\label{tab:Sensitivity}
\end{table}

The results in Table \ref{tab:Sensitivity} yield two salient observations. Firstly, DCL$_0$ demonstrates remarkable performance, surpassing the best heuristic benchmarks on perishable inventory systems and inventory systems with stochastic lead times. It also exhibits better performance on average relative to BSP in lost sales inventory control. (When analyzing instances individually, we notice that it is sometimes outperformed by BSP when the lead time is large.) Interestingly, DCL$_0$ manages to achieve optimality gaps around $1\%$, which, when placed in comparison with the prior DRL application, A3C \citep{gijsbrechts2019can}, represents a significant enhancement. To offer perspective, A3C records optimality gaps within $3-6\%$ for lost sales inventory control. This justifies our initial motivation that inventory management problems may require evaluating states under multiple exogenous scenarios for subsequent policy updates. As a result, even without the proposed techniques in the simulation part of DCL, it is possible to obtain better results against BSP (Lost Sales) and against the best heuristics (Perishable and Random Lead Times).

Furthermore, a second critical observation is that the integration of SH with CRN into the DCL algorithm yields an improvement in performance over DCL$_0$ in all cases. This integration results in optimality gaps that are reduced by more than an order of magnitude, all while operating within the same simulation resource constraints ($M=1000$). This twofold observation underscores both our initial motivation to design a DRL algorithm for problems modeled by MDP-EI and the pivotal role of the simulation techniques in increasing its efficiency.

These results prompt an important follow-up question: how many additional simulation resources would be required for DCL$_0$ to match the performance of DCL? To investigate this, we conduct a series of simulations by adjusting the number of exogenous scenarios ($M$) from the set $\{100, 500, 1000, 2000, 5000, 10000, 20000\}$, while maintaining other hyperparameters constant. (We note that the total budget $B_\s$ of exogenous scenarios for a sampled state $\s$ is $B_\s=\A_s|M|$) Since the purpose of simulations is to identify the simulation-based action $\hat{\pi}^+(\s)$ that closely approximates or matches the optimal action $\pi^+(\s)$, we examine how varying resource allocation strategies impact the accuracy of determining simulation-based actions. (We also note that the effect of M on computational time of simulations can be considered linear, for example, $M=1000$ takes takes twice as long as $M=500$.) To this end, we evaluate four simulation strategies: SH with CRN (DCL), SH without CRN, Uniform Allocation with CRN, and Uniform Allocation without CRN (DCL$_0$).
\begin{figure}[ht]
    \centering
    \includegraphics[scale=0.46]{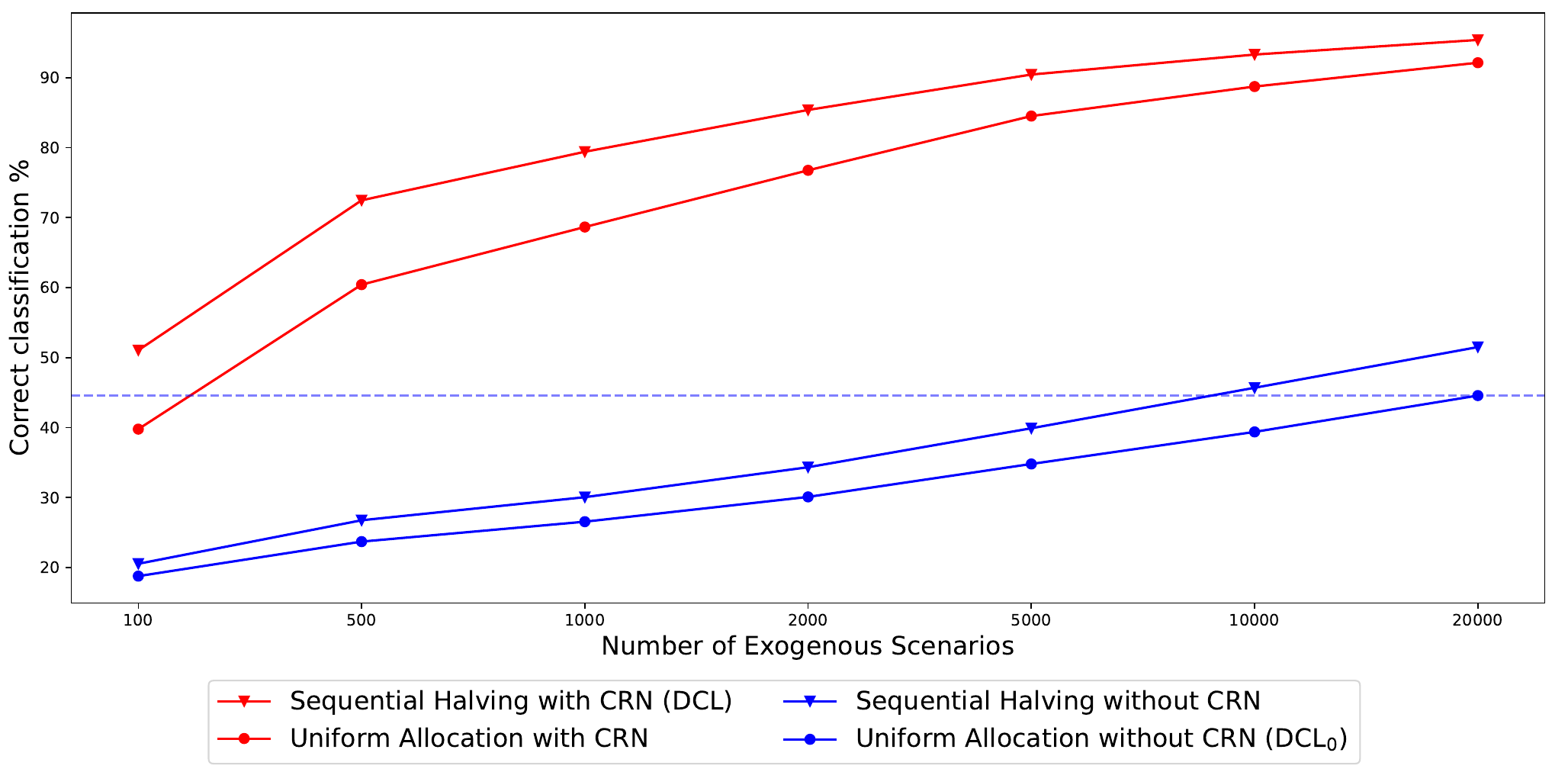}
    \caption{Correct classification \% of simulation strategies in determining the optimal actions. For $M=1000$, we highlight the accuracy results of DCL and DCL${_0}$ used in the numerical experiments. Table \ref{tab:Sensitivity} provides insights into how these accuracy results reflect the actual performance of the corresponding policies.}
        \begin{tikzpicture}[overlay, remember picture]
        \draw[red, thick] (-2.02, 9.1) circle (0.20cm);
        \node[red] at (-2.01, 9.6) {DCL}; 

        \draw[blue, thick] (-2.02, 5.3) circle (0.20cm);
        \node[blue] at (-2.01, 4.8) {DCL$_0$};
    \end{tikzpicture}\label{fig:simulationanalysis}
\end{figure}
\vspace{-5pt}

To obtain rigorous analysis, we select six diverse "small" instances from the studied inventory management problems (from lost sales inventory control with 1-) Poisson demand, $\tau=3$, $p=39$, 2-) Poisson demand, $\tau=4$, $p=19$ and 3-) geometric demand, $\tau=3$, $p=19$; from perishable inventory systems with 4-) $m_l=4$, $\tau=1$, $f=1.0$, $cvr=1.5$, 5-) $m_l=4$, $\tau=1$, $f=0.0$, $cvr=1.5$ and 6-) $m_l=3$, $\tau=1$, $f=0.5$, $cvr=1.5$), where we can computationally find the optimal action $\pi^+(\s)$ for a state $\s$ before following the policy $\pi$. Consequently, we can analyze the performance of the selected simulation algorithms by measuring the accuracy (correct classification \%) of how well they can find $\hat{\pi}^+(\s)$ such that $\hat{\pi}^+(\s)=\pi^+(\s)$ for the sampled states. 

Figure \ref{fig:simulationanalysis} provides the average results of these experiments, where $M=1000$ denotes the setting we use to obtain numerical results. We observe that the main performance gain is obtained by the inclusion of CRN when generating exogenous scenarios, hinting at a huge variance reduction when comparing the estimated expected costs of two actions. Moreover, SH provides up to $10\%$ better correct classification in pursuing the optimal action when equipped with CRN. Notably, without CRN and SH, the simulation process suffers from variance and cannot allocate resources to promising actions. To illustrate this claim, the dashed horizontal line in Figure \ref{fig:simulationanalysis} shows that DCL$_0$ would need to generate exogenous scenarios at a volume more than a hundred times higher to match the performance of DCL in finding the simulation-based actions. 

The results presented in Figure \ref{fig:simulationanalysis} show that DCL$_0$ achieves poor classification performance, which may seem at odds with the conclusions drawn from Table \ref{tab:Sensitivity}, where DCL$_0$ outperforms the heuristics. This apparent contradiction can be explained by the training process of DCL$_0$, which involves testing samples under multiple exogenous scenarios. Despite its suboptimal classification, DCL$_0$ generates simulation-based actions that remain effective, even if not always optimal. As a result, DCL$_0$ is capable of delivering commendable outcomes. Nevertheless, accurate classification contributes to enhanced performance, as demonstrated by the significant differences observed between DCL$_0$ and DCL in lost sales inventory control.

\section{Conclusion}\label{sec:discussion}

The current body of literature acknowledges the potential of DRL algorithms in addressing complex inventory management problems \citep{boute2021deep}. While the initial outcomes have been encouraging, there remains a degree of uncertainty regarding the practical application of such algorithms, particularly their ability to supersede established heuristics.

In this research, we bridge this gap by enhancing the applicability and performance of DRL in the challenging domain of inventory management. We develop an algorithm that surpasses the limitations of existing DRL applications. In particular, we introduce the DCL algorithm, tailored to address inventory management challenges. DCL outperformed top heuristics in lost sales inventory control, perishable inventory systems, and inventory systems with random lead-time. The techniques used, notably SH with CRN (referenced in \S\ref{sec:Sensitivity}), enhance simulation efficiency, elevating the performance of DCL. With consistent hyperparameters throughout our experiments, the robustness of DCL is underscored. These results substantiate our initial hypothesis, suggesting that DRL indeed holds promise for inventory management problems but requires tailored algorithms to meet their requirements. 

However, inventory management often requires balancing optimality and simplicity. While DCL significantly reduces the optimality gap compared to base-stock policies, it introduces greater complexity. Since many inventory systems are increasingly automated, this complexity could become less of a barrier, as replenishment decisions are typically managed by software. 

Looking ahead, future work could focus on two main directions. Firstly, beyond inventory problems, the methodology presented here might be transferable to other settings characterized by high stochasticity, where large numbers of exogenous scenarios can be generated to evaluate decision-making actions. Secondly, exploring the scalability of the DCL algorithm is a promising avenue. Specifically, investigating methodologies that enable DCL to generalize across multiple product lines could potentially reduce the need to train distinct models for each individual item. These advancements could significantly extend the applicability of DCL by providing a more practicable solution for complex inventory management systems.

\section*{Acknowledgements}
The authors thank the editor and the three anonymous reviewers for their valuable comments and suggestions, which have significantly improved this paper. Tarkan Temizöz conducted his research in the project DynaPlex: Deep Reinforcement Learning for Data-Driven Logistics, made possible by TKI Dinalog and the Topsector Logistics and funded by the Ministry of Economic Affairs and Climate Policy. We acknowledge the support of the SURF Cooperative using grant no. EINF-5192. 

\bibliographystyle{apalike}
\bibliography{sample}

\begin{thebibliography}{}

\bibitem[Adan et~al., 1995]{adan1995}
Adan, I., van Eenige, M., and Resing, J. (1995).
\newblock Fitting discrete distributions on the first two moments.
\newblock {\em Probability in the Engineering and Informational Sciences},
  9(4):623–632.

\bibitem[Ang et~al., 2017]{Ang2017}
Ang, M., Sigman, K., Song, J.-S., and Zhang, H. (2017).
\newblock Closed-form approximations for optimal (r, q) and (s, t) policies in
  a parallel processing environment.
\newblock {\em Operations Research}, 65(5):1414--1428.

\bibitem[Arrow and Karlin, 1958]{Arrow1958}
Arrow, K. and Karlin, S. (1958).
\newblock {\em Studies in the Mathematical Theory of Inventory and Production}.
\newblock Stanford University, Stanford, CA.

\bibitem[Bijvank et~al., 2023]{Bijvank2023}
Bijvank, M., Huh, W., and Jarakiraman, G. (2023).
\newblock Lost-sales inventory systems.
\newblock In Song, J.-S., editor, {\em Research Handbook on Inventory
  Management}, chapter~1. Edward Elgar Publishing, Cheltenham, UK.

\bibitem[Boute et~al., 2022]{boute2021deep}
Boute, R.~N., Gijsbrechts, J., {van Jaarsveld}, W., and Vanvuchelen, N. (2022).
\newblock Deep reinforcement learning for inventory control: A roadmap.
\newblock {\em European Journal of Operational Research}, 298(2):401--412.

\bibitem[Bradley and Robinson, 2005]{Bradley2005}
Bradley, J.~R. and Robinson, L.~W. (2005).
\newblock Improved base-stock approximations for independent stochastic lead
  times with order crossover.
\newblock {\em Manufacturing \& Service Operations Management}, 7(4):319--329.

\bibitem[Broekmeulen and van Donselaar, 2009]{broekmeulen2009heuristic}
Broekmeulen, R.~A. and van Donselaar, K.~H. (2009).
\newblock A heuristic to manage perishable inventory with batch ordering,
  positive lead-times, and time-varying demand.
\newblock {\em Computers \& Operations Research}, 36(11):3013--3018.

\bibitem[Bu et~al., 2023]{bu2023}
Bu, J., Gong, X., and Chao, X. (2023).
\newblock Asymptotic optimality of base-stock policies for perishable inventory
  systems.
\newblock {\em Management Science}, 69(2):846--864.

\bibitem[Bubeck and Cesa-Bianchi, 2012]{Bubeck2012}
Bubeck, S. and Cesa-Bianchi, N. (2012).
\newblock Regret analysis of stochastic and nonstochastic multi-armed bandit
  problems.
\newblock {\em Foundations and Trends® in Machine Learning}, 5(1):1--122.

\bibitem[Cachon et~al., 2020]{Cachon2020}
Cachon, G.~P., Girotra, K., and Netessine, S. (2020).
\newblock Interesting, important, and impactful operations management.
\newblock {\em Manufacturing \& Service Operations Management}, 22(1):214--222.

\bibitem[Chao et~al., 2018]{Chao2018}
Chao, X., Gong, X., Shi, C., Yang, C., Zhang, H., and Zhou, S.~X. (2018).
\newblock Approximation algorithms for capacitated perishable inventory systems
  with positive lead times.
\newblock {\em Management Science}, 64(11):5038--5061.

\bibitem[Chen et~al., 2014]{Chen2014adp}
Chen, W., Dawande, M., and Janakiraman, G. (2014).
\newblock Fixed-dimensional stochastic dynamic programs: An approximation
  scheme and an inventory application.
\newblock {\em Operations Research}, 62(1):81--103.

\bibitem[Cohen and Pekelman, 1978]{cohen1978lifo}
Cohen, M. and Pekelman, D. (1978).
\newblock Lifo inventory systems.
\newblock {\em Management Science}, 24(11):1150--1162.

\bibitem[Danihelka et~al., 2022]{danihelka2022policy}
Danihelka, I., Guez, A., Schrittwieser, J., and Silver, D. (2022).
\newblock Policy improvement by planning with gumbel.
\newblock In {\em International Conference on Learning Representations}.

\bibitem[{De Moor} et~al., 2022]{bram2022}
{De Moor}, B.~J., Gijsbrechts, J., and Boute, R.~N. (2022).
\newblock Reward shaping to improve the performance of deep reinforcement
  learning in perishable inventory management.
\newblock {\em European Journal of Operational Research}, 301(2):535--545.

\bibitem[Dehaybe et~al., 2024]{DEHAYBE2024433}
Dehaybe, H., Catanzaro, D., and Chevalier, P. (2024).
\newblock Deep reinforcement learning for inventory optimization with
  non-stationary uncertain demand.
\newblock {\em European Journal of Operational Research}, 314(2):433--445.

\bibitem[Dietterich et~al., 2018]{dietterich2018discovering}
Dietterich, T.~G., Trimponias, G., and Chen, Z. (2018).
\newblock Discovering and removing exogenous state variables and rewards for
  reinforcement learning.
\newblock In {\em International Conference on Machine Learning}, pages
  1262--1270. PMLR.

\bibitem[Disney et~al., 2016]{Disney2016}
Disney, S.~M., Maltz, A., Wang, X., and Warburton, R.~D. (2016).
\newblock Inventory management for stochastic lead times with order crossovers.
\newblock {\em European Journal of Operational Research}, 248(2):473--486.

\bibitem[Ernst et~al., 2005]{Ernst2005}
Ernst, D., Geurts, P., and Wehenkel, L. (2005).
\newblock Tree-based batch mode reinforcement learning.
\newblock {\em Journal of Machine Learning Research}, 6(18):503--556.

\bibitem[Fabiano and Cazenave, 2022]{seqhalvingscores}
Fabiano, N. and Cazenave, T. (2022).
\newblock Sequential halving using scores.
\newblock In {\em Advances in Computer Games}, pages 41--52, Cham. Springer
  International Publishing.

\bibitem[Francesco~Stranieri and Kouki, 2024]{Stranieri2024}
Francesco~Stranieri, F.~S. and Kouki, C. (2024).
\newblock Performance of deep reinforcement learning algorithms in two-echelon
  inventory control systems.
\newblock {\em International Journal of Production Research},
  62(17):6211--6226.

\bibitem[Geevers et~al., 2023]{geevers2023multiechelon}
Geevers, K., van Hezewijk, L., and Mes, M.~R. (2023).
\newblock Multi-echelon inventory optimization using deep reinforcement
  learning.
\newblock Available at SSRN: http://dx.doi.org/10.2139/ssrn.4227665.

\bibitem[Gijsbrechts et~al., 2022]{gijsbrechts2019can}
Gijsbrechts, J., Boute, R.~N., Van~Mieghem, J.~A., and Zhang, D.~J. (2022).
\newblock Can deep reinforcement learning improve inventory management?
  performance on lost sales, dual-sourcing, and multi-echelon problems.
\newblock {\em Manufacturing \& Service Operations Management},
  24(3):1349--1368.

\bibitem[Goldberg et~al., 2016]{goldberg2016asymptotic}
Goldberg, D.~A., Katz-Rogozhnikov, D.~A., Lu, Y., Sharma, M., and Squillante,
  M.~S. (2016).
\newblock Asymptotic optimality of constant-order policies for lost sales
  inventory models with large lead times.
\newblock {\em Mathematics of Operations Research}, 41(3):898--913.

\bibitem[Goodfellow et~al., 2016]{goodfellow2016deep}
Goodfellow, I., Bengio, Y., and Courville, A. (2016).
\newblock {\em Deep Learning}.
\newblock MIT Press.
\newblock \url{http://www.deeplearningbook.org}.

\bibitem[Haijema and Minner, 2019]{Haijema2019}
Haijema, R. and Minner, S. (2019).
\newblock Improved ordering of perishables: The value of stock-age information.
\newblock {\em International Journal of Production Economics}, 209:316--324.
\newblock The Proceedings of the 19th International Symposium on Inventories.

\bibitem[Hong et~al., 2021]{hong2021review}
Hong, L.~J., Fan, W., and Luo, J. (2021).
\newblock Review on ranking and selection: A new perspective.
\newblock {\em Frontiers of Engineering Management}, 8:321--343.

\bibitem[Huh et~al., 2009]{huh2009asymptotic}
Huh, W.~T., Janakiraman, G., Muckstadt, J.~A., and Rusmevichientong, P. (2009).
\newblock Asymptotic optimality of order-up-to policies in lost sales inventory
  systems.
\newblock {\em Management Science}, 55(3):404--420.

\bibitem[Johansen and Thorstenson, 2008]{Johansen2008}
Johansen, S. and Thorstenson, A. (2008).
\newblock Pure and restricted base-stock policies for the lost-sales inventory
  system with periodic review and constant lead times.
\newblock In {\em 15th International Symposium on Inventories}.

\bibitem[Karaesmen et~al., 2011]{karaesmen2011managing}
Karaesmen, I., Scheller-Wolf, A., and Deniz, B. (2011).
\newblock Managing perishable and aging inventories: Review and future research
  directions.
\newblock In {\em Planning Production and Inventories in the Extended
  Enterprise}, volume 151 of {\em International Series in Operations Research
  \& Management Science}, pages 393--436. Springer, New York.

\bibitem[Karnin et~al., 2013]{seqhalving}
Karnin, Z., Koren, T., and Somekh, O. (2013).
\newblock Almost optimal exploration in multi-armed bandits.
\newblock In {\em Proceedings of the 30th International Conference on Machine
  Learning}, volume 28(3) of {\em Proceedings of Machine Learning Research},
  pages 1238--1246.

\bibitem[Kaynov et~al., 2024]{KAYNOV2024109088}
Kaynov, I., {van Knippenberg}, M., Menkovski, V., {van Breemen}, A., and {van
  Jaarsveld}, W. (2024).
\newblock Deep reinforcement learning for one-warehouse multi-retailer
  inventory management.
\newblock {\em International Journal of Production Economics}, 267:109088.

\bibitem[Lagoudakis and Parr, 2003]{lagoudakis2003reinforcement}
Lagoudakis, M.~G. and Parr, R. (2003).
\newblock Reinforcement learning as classification: Leveraging modern
  classifiers.
\newblock In {\em Proceedings of the 20th International Conference on Machine
  Learning (ICML-03)}, pages 424--431.

\bibitem[Law and Kelton, 2000]{law2000simulation}
Law, A.~M. and Kelton, W.~D. (2000).
\newblock {\em Simulation Modeling and Analysis}.
\newblock McGraw-Hill, Boston, MA.

\bibitem[Lazaric et~al., 2016]{Lazaric}
Lazaric, A., Ghavamzadeh, M., and Munos, R. (2016).
\newblock Analysis of classification-based policy iteration algorithms.
\newblock {\em Journal of Machine Learning Research}, 17(19):1--30.

\bibitem[Mao et~al., 2019]{mao2018variance}
Mao, H., Venkatakrishnan, S.~B., Schwarzkopf, M., and Alizadeh, M. (2019).
\newblock Variance reduction for reinforcement learning in input-driven
  environments.
\newblock In {\em International Conference on Learning Representations}.

\bibitem[Minner and Transchel, 2010]{minner2010}
Minner, S. and Transchel, S. (2010).
\newblock Periodic review inventory-control for perishable products under
  service-level constraints.
\newblock {\em OR Spectrum}, 32:979--996.

\bibitem[Mnih et~al., 2016]{mnih2016asynchronous}
Mnih, V., Badia, A.~P., Mirza, M., Graves, A., Lillicrap, T., Harley, T.,
  Silver, D., and Kavukcuoglu, K. (2016).
\newblock Asynchronous methods for deep reinforcement learning.
\newblock In {\em International conference on machine learning}, pages
  1928--1937.

\bibitem[Mnih et~al., 2015]{mnih2015human}
Mnih, V., Kavukcuoglu, K., Silver, D., Rusu, A.~A., Veness, J., Bellemare,
  M.~G., Graves, A., Riedmiller, M., Fidjeland, A.~K., Ostrovski, G., et~al.
  (2015).
\newblock Human-level control through deep reinforcement learning.
\newblock {\em nature}, 518(7540):529--533.

\bibitem[Morton, 1969]{morton1969bounds}
Morton, K. (1969).
\newblock Bounds on the solution of the lagged optimal inventory equation with
  no demand backlogging and proportional costs.
\newblock {\em SIAM Review}, 11(4):572--596.

\bibitem[Morton, 1971]{morton1971near}
Morton, K. (1971).
\newblock The near-myopic nature of the lagged-proportional-cost inventory
  problem with lost sales.
\newblock {\em Operations Research}, 19(7):1708--1716.

\bibitem[Muthuraman et~al., 2015]{Muthuraman2015}
Muthuraman, K., Seshadri, S., and Wu, Q. (2015).
\newblock Inventory management with stochastic lead times.
\newblock {\em Mathematics of Operations Research}, 40(2):302--327.

\bibitem[Nahmias, 1975a]{nahmias1975comparison}
Nahmias, S. (1975a).
\newblock A comparison of alternative approximations for ordering perishable
  inventory.
\newblock {\em Information Systems and Operational Research}, 13(2):175--184.

\bibitem[Nahmias, 1975b]{nahmias1975optimal}
Nahmias, S. (1975b).
\newblock Optimal ordering policies for perishable inventory—ii.
\newblock {\em Operations Research}, 23(4):735--749.

\bibitem[Nahmias, 2011]{nahmias2011perishable}
Nahmias, S. (2011).
\newblock {\em Perishable Inventory Systems}.
\newblock International Series in Operations Research \& Management Science.
  Springer, New York.

\bibitem[Oroojlooyjadid et~al., 2021]{oroojlooyjadid2021deep}
Oroojlooyjadid, A., Nazari, M., Snyder, L.~V., and Tak{\'a}{\v{c}}, M. (2021).
\newblock A deep q-network for the beer game: Deep reinforcement learning for
  inventory optimization.
\newblock {\em Manufacturing \& Service Operations Management}, 24(1):285--304.

\bibitem[Powell, 2020]{powell2020reinforcement}
Powell, W. (2020).
\newblock {\em Reinforcement Learning and Stochastic Optimization: A unified
  framework for sequential decisions}.
\newblock Wiley-Interscience.

\bibitem[Powell, 2011]{powell2011approximate}
Powell, W.~B. (2011).
\newblock {\em Approximate Dynamic Programming: Solving the Curses of
  Dimensionality}.
\newblock Wiley, Hoboken, NJ.

\bibitem[Powell, 2019]{powell2019unified}
Powell, W.~B. (2019).
\newblock A unified framework for stochastic optimization.
\newblock {\em European Journal of Operational Research}, 275(3):795--821.

\bibitem[Puterman, 2014]{puterman2014markov}
Puterman, M.~L. (2014).
\newblock {\em Markov decision processes: discrete stochastic dynamic
  programming}.
\newblock John Wiley \& Sons.

\bibitem[Schulman et~al., 2017]{schulman2017proximal}
Schulman, J., Wolski, F., Dhariwal, P., Radford, A., and Klimov, O. (2017).
\newblock Proximal policy optimization algorithms.
\newblock {\em arXiv preprint arXiv:1707.06347}.

\bibitem[Silver et~al., 2018]{silver2018general}
Silver, D., Hubert, T., Schrittwieser, J., Antonoglou, I., Lai, M., Guez, A.,
  Lanctot, M., Sifre, L., Kumaran, D., Graepel, T., et~al. (2018).
\newblock A general reinforcement learning algorithm that masters chess, shogi,
  and go through self-play.
\newblock {\em Science}, 362(6419):1140--1144.

\bibitem[Silver et~al., 2016]{Silver2016}
Silver, E., Pyke, D., and Thomas, D. (2016).
\newblock Chapter 1.
\newblock In {\em Inventory and Production Management in Supply Chains}. CRC
  Press, 4th edition.

\bibitem[Stolyar and Wang, 2022]{Stolyar2022}
Stolyar, A.~L. and Wang, Q. (2022).
\newblock Exploiting random lead times for significant inventory cost savings.
\newblock {\em Operations Research}, 70(4):2496--2516.

\bibitem[Sun et~al., 2016]{sun2016quadratic}
Sun, P., Wang, K., and Zipkin, P. (2016).
\newblock Quadratic approximation of cost functions in lost sales and
  perishable inventory control problems.
\newblock Working paper, Fuqua School of Business, Duke University, Durham, NC.

\bibitem[Sutton and Barto, 2018]{sutton2018reinforcement}
Sutton, R.~S. and Barto, A.~G. (2018).
\newblock {\em Reinforcement Learning: An Introduction}.
\newblock MIT Press, 2nd edition.

\bibitem[Tesauro and Galperin, 1996]{Tesauro}
Tesauro, G. and Galperin, G. (1996).
\newblock On-line policy improvement using monte-carlo search.
\newblock In {\em Advances in Neural Information Processing Systems}, volume~9.
  MIT Press.

\bibitem[Trimponias and Dietterich, 2023]{trimponias2023reinforcement}
Trimponias, G. and Dietterich, T.~G. (2023).
\newblock Reinforcement learning with exogenous states and rewards.
\newblock {\em arXiv preprint arXiv:2303.12957}.

\bibitem[van Hezewijk et~al., 2023]{vanHezewijk2023}
van Hezewijk, L., Dellaert, N.~P., van Woensel, T., and Gademann, A. J. R.~M.
  (2023).
\newblock Using the proximal policy optimisation algorithm for solving the
  stochastic capacitated lot sizing problem.
\newblock {\em International Journal of Production Research}, 61(6):1955--1978.

\bibitem[Van~Houtum and Kranenburg, 2015]{Houtum2015spare}
Van~Houtum, G.-J. and Kranenburg, B. (2015).
\newblock {\em Spare parts inventory control under system availability
  constraints}, volume 227.
\newblock Springer.

\bibitem[Vanvuchelen et~al., 2020]{vanvuchelenPPO}
Vanvuchelen, N., Gijsbrechts, J., and Boute, R. (2020).
\newblock {Use of Proximal Policy Optimization for the Joint Replenishment
  Problem}.
\newblock {\em Computers in Industry}, 119:103239.

\bibitem[Wang and Minner, 2024]{WANG2024109133}
Wang, Y. and Minner, S. (2024).
\newblock Deep reinforcement learning for demand fulfillment in online retail.
\newblock {\em International Journal of Production Economics}, 269:109133.

\bibitem[Xin, 2021]{Xin2021}
Xin, L. (2021).
\newblock Technical note—understanding the performance of capped base-stock
  policies in lost-sales inventory models.
\newblock {\em Operations Research}, 69(1):61--70.

\bibitem[Xin and Goldberg, 2016]{xin2016optimality}
Xin, L. and Goldberg, D.~A. (2016).
\newblock Optimality gap of constant-order policies decays exponentially in the
  lead time for lost sales models.
\newblock {\em Operations Research}, 64(6):1556--1565.

\bibitem[Zipkin, 2008]{zipkin2008old}
Zipkin, P. (2008).
\newblock Old and new methods for lost-sales inventory systems.
\newblock {\em Operations Research}, 56(5):1256--1263.

\bibitem[Zipkin, 2000]{zipkin2000foundations}
Zipkin, P.~H. (2000).
\newblock {\em Foundations of Inventory Management}.
\newblock McGraw-Hill, Boston.

\end{thebibliography}
\newpage
\appendix

\section{Formulating the inventory problems as MDP-EI}\label{app:invmodelformulation}

Here we characterize the inventory problems as MDP-EI models $\mathcal{M} = \langle \mathcal{S}, \A, \mathcal{W}, \Input, f, C, \alpha, \s_{0} \rangle$. We consider a single product and adopt a finite action set $\A = \{0,1,\dots,m\}$, where actions represent the amount ordered by the decision-maker. Consistent with the literature, we aim to minimize the average cost over an infinite horizon and hence $\alpha = 1$. The algorithm is rather insensitive to $\s_0$; we let the initial state correspond to having no stock on hand or in the pipeline. 

\noindent \textbf{Lost sales inventory control}

We consider the well-known discrete-time lost sales inventory system \citep[][]{zipkin2008old}. In this system, we manage an inventory of items over time by placing orders for new items, attempting to meet the exogenous demands, and incurring costs based on whether we have sufficient inventory to meet the demand or not (in case of lost sales, partial fills are assumed). The constant lead time for new orders to arrive is greater than one period (i.e., $\tau>1$).

In this setting, states $\s$ are represented as vectors in $\mathbb{R}^{\tau}$, where each element of the vector represents a component of the inventory: $\s[1]$ corresponds to on-hand inventory or the current availability of items, and $\s[2]$ to $\s[\tau]$ depict pipeline inventory, which are the items ordered and due to arrive in $1,\ldots,\tau-1$ future periods, respectively. For this system, the per-period demand, denoted by $D$, is the distribution governing the random variable $\mathbf{\Xi}$, where $\xi_t \sim D$ represents a realization of the demand at each time step $t$, and the exogenous input space is defined as $\mathcal{W} := \mathbb{Z}_0$. The transition function $f(\s,a,\xi)=\s'$ describes the evolution of the state when an action $a$ is taken in state $\s$ under demand $\xi$. The on-hand inventory is updated by meeting the demand with the current on-hand inventory and then adding any pipeline inventory due to arrive, leading to $\s'[1]=(\s[1]-\xi)^+ + \s[2]$. Action $a$ corresponds to the order due to arrive in $\tau$ time periods, hence $\s'[\tau]=a$. For all $i\in \{2,\ldots,\tau-1\}$, pipeline inventory is shifted one period closer to arrival, with $\s'[i]=\s[i+1]$. The transition function is therefore given as $f((\s[1],\ldots,\s[\tau]),a,\xi):=((\s[1]-\xi)^++\s[2], \s[3], \ldots,\s[\tau],a)$.

The cost function is $C(\s,a,\xi)=h(\s[1]-\xi)^+ + p(\xi-\s[1])^+$, where $h$ is the unit cost of holding inventory and $p$ is the unit penalty cost of lost sales. (In this case, action $a$ does not directly influence the cost in the current period.)

It remains to discuss $\pi_0$, $|\A|=m+1$, and $\A_s$ for each state. Let $I_{\textrm{max}}$ denote the newsvendor fractile for cumulative demand over $\tau$ periods, which bounds the optimal inventory position \cite[see][]{zipkin2008old}. Then $\pi_0$ will be a BSP with base-stock level $I_{\textrm{max}}$, and $\A_{\s}$ for state $\s$ is constructed to ensure that the inventory position does not exceed $I_{\textrm{max}}$. The maximum order quantity $m$ corresponds to the single-period newsvendor fractile bound \citep[see][]{zipkin2008old}.

\noindent \textbf{Perishable inventory systems}

We consider a periodic review perishable inventory system of a product possessing a maximum life of $m_l$ periods and having a lead time $\tau \geq 0$, following \citet{Haijema2019}. Accordingly, the exogenous input encapsulates two stochastic variables $\xi=(\xi^{FIFO}, \xi^{LIFO})$ with $\xi \in \mathcal{W} := \mathbb{Z}_0$, which represent the number of demands observed at the end of a period under FIFO and LIFO issuance policies, respectively.

States $\s$ are depicted as vectors in $\R^{m_l+max(\tau-1,0)}$, where $\s[i]$, for $i=1,\dots,m_l$, denotes the inventory level of items with at $i$ period lifetime remaining. If the lead time $\tau > 1$, $\s[i]$, for $i=m_l+1,\dots,m_l+\tau-1$, denotes the amount of pipeline inventory arriving in the next $i - m_l$ periods. We refer to \citet{Haijema2019} for a complete description of state transitions. We denote the cost function as $C(\s,a,\xi) = wn_{\textrm{perish}}+h(OH-d-n_{\textrm{perish}})^++p(d-OH)^+$, where $n_{\textrm{perish}}$ denotes the number of perished units at the end of a period, $OH$ denotes the amount of on-hand inventory at the end of a period, $w$ denotes the waste cost of a unit perished, $h$ denotes holding cost, $p$ denotes penalty cost, and $d=\xi^{FIFO}+\xi^{LIFO}$ represents the total demand.

We will employ BSP as the initial policy $\pi_0$. We limit the maximum inventory position by $I_{max}= m$, equating these parameters to the newsvendor solution required to cater to demand over $\tau + m_l$ periods, i.e., when the inventory on hand and in the pipeline have turned into waste \citep[see][]{Haijema2019}.

\noindent \textbf{Inventory systems with random lead times}

We consider a continuous review inventory system with independent, identically distributed replenishment lead times with order-crossing and backlogged unsatisfied demands. Crucially, the decision-maker lacks access to order arrival times for orders in the pipeline. To simplify the exposition, we assume a fixed maximum for the allowable pipeline inventory, denoted as $il_{max}$.

States will be represented as a vector $\s=(\s[1],\s[2],\x)$, where $\x$ represents a vector of length $\R^{il_{max}}$. Here, $\s[1]$ signifies the on-hand inventory, $\s[2]$ indicates the total pipeline inventory, and the elements $\x[i], \text{ } i\in \{1,\ldots,\s[2]\}$ with $\s[2]>0$, denote the elapsed time since the placement of each yet-to-be-received order, while $\x[i] = 0$ for $i>\s[2]$. When $\s[2]=0$, $\x[i] = 0, \text{ } i\in \{1,\ldots,il_{max}\}$. If the lead time is exponentially distributed, information regarding elapsed time for pending orders becomes irrelevant due to the memoryless nature of the exponential distribution. As a result, a state $\mathbf{s}$ corresponds to a vector in $\R^2$ that represents the inventory level and the inventory in the pipeline. This reduction facilitates the computation of the optimal policy. Since we consider a continuous review, continuous time inventory system with several stochastic inputs, i.e., demand and lead times, we employ several strategies to formulate this problem as MDP-EI while staying close to \citet{Stolyar2022}. 

We first explain how to construct the exogenous inputs in inventory systems with random lead times and then describe the state and the cost transitions. We define a (decision) epoch in this system by the duration between two consecutive demands. For each epoch, the exogenous input $\xi$ is depicted as a vector in $\R^{il_{max}+1}$. The element $\xi[il_{max}+1]$ denotes the interarrival time between the current demand and the next one, i.e., between the current decision epoch and the next. The elements $\xi[i]$, where $i=1,\dots, il_{max}$, determine for each order in the pipeline (including those orders placed in the current epoch) whether it will arrive before the next epoch and when.  

It is sufficient to let $\xi[il_{max}+1]$ be distributed following the interarrival time of demand, while $\xi[i]$ for $i=1, \dots, il_{max}$ will be uniformly distributed, regardless of the lead-time distribution. In particular, after placing (zero or more) orders in a decision epoch, $\xi$ is observed, and the state of the system at the beginning of the next epoch can be determined as follows. Consider any order $i$ ($i\in \{1,\ldots,\s[2]\}$) that is in the pipeline at the start of the current epoch (after placing orders). It will arrive before the next epoch with probability $p=\mathbf{P}(L\le\x[i]+\xi[il_{max}+1]|L>\x[i])$, where we denote by $L$ a generic random variable representing a replenishment lead time. Now, if $\xi[i]\le p$, then we interpret this as meaning that the order has arrived during the current epoch, and the inventory position $\s[1]$, the pipeline $\s[2]$, and the vector $\x$ are updated accordingly. The value of $\xi[i]$ can also be used to construct the arrival time of the order (between the time associated with the current and next epoch). The arrival times can be used to derive an expression for the cost function $C(\s,a,\xi)$, i.e., the holding costs and backorder costs accrued between the current and the next epoch. (This expression is a bit involved but mostly straightforward and therefore omitted.) If $\xi[i]>p$, the order is still in the pipeline at the start of the next epoch, and $\x[i]$ is updated accordingly. Specifically, when the next epoch starts, the time elapsed since $i$ was placed equals $\x[i]+\xi[il_{max}+1]$. In the next epoch, it will again be determined whether the order will arrive. After reviewing each order, the demand is observed, and the holding costs and backorder costs are incurred considering the time between the last incoming order and the demand (or the interarrival time of the demand ($\xi[il_{max}+1]$) if no order is received). Then the inventory level is decreased by one, and the next epoch starts.  

We will use BSP as the initial policy $\pi_0$. Drawing from the approach detailed by \citet{Stolyar2022} (see the e-companion for the corresponding calculations), we truncate the infinite state space to apply DCL feasibly, setting specific boundaries for this purpose. Notably, the decision-maker maintains the inventory position within a fixed, finite range. This strategy is reflected in the inequality $I_{min} \leq \s[1]+\s[2]+a \leq I_{max}$, where $I_{min}$ and $I_{max}$ denote the minimum and maximum inventory levels, respectively. We also implement a cap on the maximum number of back-orders, denoted as $BO_{max}$, and any unsatisfied demand after this limit is lost, incurring the same backorder cost. Consequently, the maximum number of orders that can be in the pipeline inventory can be computed as $il_{max} = I_{max} + BO_{max}$. We note that these boundaries only truncate those states that are less likely to be visited under any policy, allowing us to focus on the more probable and relevant states. We set the maximum order quantity $m$ high enough such that a well-performing policy, GBS policy by \citet{Stolyar2022}), is likely to order less than this amount.

\section{Examples}\label{app:examples}

\begin{example}
\label{ex1:api} The effect of multiple exogenous scenarios on action-value estimates.

\noindent Recall the aforementioned lost sales inventory model, formulated as an MDP-EI model in Appendix \ref{app:invmodelformulation}. Consider a system with lead time $\tau=2$, where the applicable actions are $a \in \A =\{0, 1\}$, and in each period, and we observe demand $\xi \in \mathcal{W} = \{0, 1\}$. We define the per unit holding cost $h=1$, and per unit penalty cost $p=9$. We adopt the average-cost criterion, i.e., $\alpha=1.0$.  Initially, we have one unit of on-hand inventory and no unit in the pipeline, $\s_0=(1,0)$. In this setting, we start with an initial policy $\pi$, which consistently places an order of $1$ unit irrespective of the current state. Our objective is to examine how this policy can be improved for state $\s_0$ by estimating the action-value function of both actions, $a=0$ and $a=1$. Table \ref{tab:exampleqapprox} shows the approximate trajectory costs of $a=0,1$ under three generated exogenous scenarios with the horizon length $H=4$. We approximate the trajectory costs as $\hat{Q}_\pi(\s,a|\boldsymbol\xi) = \sum_{t=0}^3 c_t$. While for scenarios $\boldsymbol\xi_1$ and $\boldsymbol\xi_2$, action $a=0$ yields lower approximate trajectory costs than action $a=1$, in scenario $\boldsymbol\xi_3$, action $a=1$ actually results in a much lower approximate trajectory cost than action $a=0$. Consequently, when taking averages of the trajectory cost over three exogenous scenarios to obtain action-value estimates for actions, we observe that action $a=1$ has a lower action-value estimate, such that $\hat{q}_{\pi}(\s,a|\s=\s_0,a=1) = 6.33$ where $\hat{q}_{\pi}(\s,a|\s=\s_0,a=0) = 8$. As demonstrated in this case, considering multiple exogenous scenarios during the policy improvement process can prevent the selection of an inferior action that may appear superior under limited scenarios. 

\begin{table}[ht]
    \centering
  \footnotesize
    \begin{tabular}{|c|c|ccccc|ccccc|ccccc|}
\cmidrule{3-17}   
\multicolumn{2}{c|}{\multirow{2}[4]{*}{}} & \multicolumn{15}{c|}{Scenarios} \\
\cmidrule{3-17}  
\multicolumn{2}{c|}{\multirow{2}[4]{*}{}} 
&  \multicolumn{5}{c|}{$\boldsymbol\xi_1 = \{0,0,0,0\}$}
&  \multicolumn{5}{c|}{$\boldsymbol\xi_2 = \{0,1,0,1\}$}
&  \multicolumn{5}{c|}{$\boldsymbol\xi_3 = \{1,1,1,1\}$}\\
\cmidrule{3-17}  
\multicolumn{2}{c|}{\multirow{2}[4]{*}{}} & $\s_t$ & $a_t$ & $\xi_t$ & $c_t$ & $\hat{Q}_\pi$ & $\s_t$ & $a_t$  & $\xi_t$ & $c_t$ & $\hat{Q}_\pi$ & $\s_t$ & $a_t$  & $\xi_t$ & $c_t$ & $\hat{Q}_\pi$\\
\midrule
\multirow{4}[2]{*}{Action $a = 0$}
& t = 0 & (1, 0) & 0 & 0 & 1 & & (1, 0) & 0 & 0 & 1 & & (1, 0) & 0 & 1 & 0 &\\
& t = 1 & (1, 0) & 1 & 0 & 1 & & (1, 0) & 1 & 1 & 0 & & (0, 0) & 1 & 1 & 9 &\\
& t = 2 & (1, 1) & 1 & 0 & 1 & & (0, 1) & 1 & 0 & 0 & & (0, 1) & 1 & 1 & 9 &\\
& t = 3 & (2, 1) & 1 & 0 & 2 &5& (1, 1) & 1 & 1 & 0 &1& (1, 1) & 1 & 1 & 0 &18\\
\midrule
\multirow{4}[2]{*}{Action $a = 1$}
& t = 0 & (1, 0) & 1 & 0 & 1 & & (1, 0) & 1 & 0 & 1 & & (1, 0) & 1 & 1 & 0 &\\
& t = 1 & (1, 1) & 1 & 0 & 1 & & (1, 1) & 1 & 1 & 0 & & (0, 1) & 1 & 1 & 9 &\\
& t = 2 & (2, 1) & 1 & 0 & 2 & & (1, 1) & 1 & 0 & 1 & & (1, 1) & 1 & 1 & 0 &\\
& t = 3 & (3, 1) & 1 & 0 & 3 &7& (2, 1) & 1 & 1 & 1 &3& (1, 1) & 1 & 1 & 0 &9\\
    \bottomrule
    \end{tabular}
       \caption{The approximated trajectory costs of actions under various exogenous scenarios. ($\hat{Q}_\pi(\s_0,a|\boldsymbol\xi)$ is displayed as $\hat{Q}_\pi$.)}
  \label{tab:exampleqapprox}
\end{table}
\end{example}

\begin{example}
\label{ex2:crn} The effect of CRN on decreasing the variance when calculating the action-value estimates.

\noindent A straightforward intuition suggests that the covariance term \eqref{eq:variance} may be positive in many inventory management problems. In such scenarios, actions typically denote the quantity ordered in a period. When comparing actions across diverse exogenous demand sequences, a positive covariance between actions $a$ and $a'$ suggests that their associated approximate trajectory costs will exhibit similar patterns under analogous demand sequences. For instance, under a sequence characterized by high demand, the corresponding trajectory costs for each action would be elevated compared to those under a sequence with low demand.

For the Example \ref{ex1:api}, CRN enable the generation of the same exogenous scenarios for both applicable actions $a=0$ and $a=1$. Given the state $\s_0$, the approximate trajectory costs for action $a=0$ are calculated as 5, 3, and 18, giving $var[\hat{Q}_\pi(\s_0,0|\boldsymbol{\xi})]=52.66$. The approximate trajectory costs for action $a=1$ are 7, 5, and 9, yielding $var[\hat{Q}_\pi(\s_0,1|\boldsymbol{\xi})]=6.22$. The realizations for the estimator $X=\hat{q}_\pi(\s_0,1)-\hat{q}_\pi(\s_0,0)$, can be found as 2, 2 and -9, resulting in $var[X]= 26.87$. Since $var[X] <  var[\hat{Q}_\pi(\s_0,0|\boldsymbol{\xi})] + var[\hat{Q}_\pi(\s_0,1|\boldsymbol{\xi})]$ ($26.87 < 52.66+6.22$), employing CRN is beneficial for this problem instance.
\end{example}

\section{Training neural network classifiers}\label{app:neuralnetworktraining}

The training of the neural network and the subsequent update of its parameters are grounded in a standard set of strategies, as delineated in the \emph{Deep Learning} book \citep{goodfellow2016deep}, and further detailed in Algorithm \ref{alg:training}. These techniques are widely applied across various classification tasks while there exist opportunities for further refinement.

\begin{algorithm}
\caption{Classifier}\label{alg:training}
\begin{algorithmic}[1]
\State \textbf{Input}: $\NN_\theta$, $\K$
\State \textbf{Initialize}: Neural networks parameters: $\theta$
\State Shuffle $\K$ and construct $\K^{Tr}$ and $\K^{Val}$
\For{$e=1, \dots, MaxEpoch $}
\State Construct mini-batches $\K^{Tr}_{mb}$; $\forall\K^{Tr}_{mb} \subset \K^{Tr}$, $|K^{Tr}_{mb}| = MiniBatchSize$
\ForAll{$\K^{Tr}_{mb} \subset \K^{Tr}$}
\State Compute $\mathcal{L}_\theta(\K^{Tr}_{mb})$ by \eqref{eq:lossfnc}
\State $\theta = Optimizer(\theta, \mathcal{L}_\theta(\K^{Tr}_{mb}))$, \citep[see][]{goodfellow2016deep}  \label{alg4:optimizer}
\EndFor
\State Compute $\mathcal{L}_\theta(\K^{Val})$ by \eqref{eq:lossfnc}
\If{$StoppingCondition(\mathcal{L}_\theta(\K^{Val})$), \citep[see][]{goodfellow2016deep}}
\State \textbf{break}
\EndIf
\State Shuffle $\K^{Tr}$
\EndFor
\State $\pi_\theta(\s) = \argmax_{a\in \A_{\s}} (\NN_{\theta}(\s)[a]), \text{ }\forall \s \in \mathcal{S}$
\State \textbf{Output}: $\pi_\theta$
\end{algorithmic}
\end{algorithm}

Algorithm \ref{alg:training} takes a dataset and a \emph{neural network structure} $\NN_\theta$, encapsulating the network architecture, training details, and hyperparameters, as inputs. The parameters of the neural network undergo updates via gradient descent, driven by the loss function computed from each mini-batch. The loss for a given sample $(\s,\hat{\pi}^+(\s))$ is defined using the \emph{cross-entropy} loss, a common loss function employed in classification tasks which gauges the discrepancy between the softmax of the output of the neural network (predicted probabilities for actions) for state $\s$ (where actions $a \notin A_\s$ are masked) and the simulation-based action $\hat{\pi}^+(\s)$. The loss function for a sample $(\s,\hat{\pi}^+(\s))$ can be thus formulated as:
\begin{align}
\l(\s,\hat{\pi}^+(\s))|\theta) := -\log \frac{\exp\big(\NN_{\theta}(\s)[{\hat{\pi}^+(\s)}]\big)}{\sum_{a'\in \A_\s}\exp\big(\NN_{\theta}(\s)[{a'}]\big)}, \label{eq:loss}
\end{align}
where $\NN_{\theta}(\s)[a]$ represents the predicted probability of the neural network for action $a$. Subsequently, the average loss function over a dataset $\K$ can be defined as:
\begin{align}
\mathcal{L}_\theta(\K) := \frac{1}{|\K|} \sum_{(\s,\hat{\pi}^+(\s)) \in \K} l(\s,\hat{\pi}^+(\s)|\theta) .\label{eq:lossfnc}
\end{align}

The neural network policy is defined as $\pi_\theta(\s)=\argmax_{a\in \A_{\s}} (\NN_{\theta}(\s)[a])$, valid for $\forall \s \in \mathcal{S}$, upon the completion of training. For the complete details of our implementation of neural networks and Algorithm \ref{alg:training}, we refer the readers to our GitHub page \href{https://github.com/tarkantemizoz/DynaPlex}{https://github.com/tarkantemizoz/DynaPlex}.

\end{document}